\documentclass[preprint,authoryear, 12pt]{elsarticle}

\usepackage{xcolor,soul}  
\soulregister\cite7
\soulregister\citep7
\soulregister\citet7
\soulregister\ref7
\soulregister\pageref7

\usepackage{amssymb}
\usepackage{amsthm}
\usepackage{amsmath}
\usepackage{multirow}
\usepackage{soul}
\usepackage{svg}
\usepackage{algorithmic}

\usepackage{threeparttable}
\usepackage[linesnumbered,ruled]{algorithm2e}
\usepackage{tikz}

\usepackage[pagewise]{lineno}
\usepackage{hyperref}
\hypersetup{
    colorlinks=true,
    linkcolor=blue,
    filecolor=blue,      
    urlcolor=blue,
    citecolor=cyan,
}

\def\TG{\textcolor{green}}
\def\TR{\textcolor{red}}

\newcommand{\norm}[1]{\left\lVert#1\right\rVert}

\def\LF{\mathcal{LF}}

\DeclareMathAlphabet\mathcalbf{OMS}{cmsy}{b}{n}

\journal{Journal of Photogrammetry and Remote Sensing}

\begin{document}

	\begin{frontmatter}

	\title{HCTO: Optimality-Aware LiDAR Inertial Odometry with Hybrid Continuous Time Optimization for Compact Wearable Mapping System}
		
        \author[ntu]{Jianping Li}
        \ead{jianping.li@ntu.edu.sg}

        \author[ntu]{Shenghai~Yuan}
        \ead{shyuan@ntu.edu.sg}

        \author[ntu]{Muqing~Cao}
        \ead{mqcao@ntu.edu}

        \author[ntu]{Thien-Minh Nguyen}
        \ead{thienminh.nguyen@ntu.edu}

        \author[ntu]{Kun Cao}
        \ead{kun001@e.ntu.edu.sg}

        \author[ntu]{Lihua Xie}
        \ead{elhxie@ntu.edu.sg}

        \address[ntu]{School of Electrical and Electronic Engineering, Nanyang Technological University, Singapore 639798}
    
        \begin{abstract}
        Compact wearable mapping system (WMS) has gained significant attention due to their convenience in various applications. Specifically, it provides an efficient way to collect prior maps for 3D structure inspection and robot-based “last-mile delivery” in complex environments. However, vibrations in human motion and the uneven distribution of point cloud features in complex environments often lead to rapid drift, which is a prevalent issue when applying existing LiDAR Inertial Odometry (LIO) methods on low-cost WMS. To address these limitations, we propose a novel LIO for WMSs based on Hybrid Continuous Time Optimization (HCTO) considering the optimality of Lidar correspondences. First, HCTO recognizes patterns in human motion (high-frequency part, low-frequency part, and constant velocity part) by analyzing raw IMU measurements. Second, HCTO constructs hybrid IMU factors according to different motion states, which enables robust and accurate estimation against vibration-induced noise in the IMU measurements. Third, the best point correspondences are selected using optimal design to achieve real-time performance and better odometry accuracy. We conduct experiments on head-mounted WMS datasets to evaluate the performance of our system, demonstrating significant advantages over state-of-the-art methods. Video recordings of experiments can be found on the project page of HCTO: \href{https://github.com/kafeiyin00/HCTO}{https://github.com/kafeiyin00/HCTO}. 
        
        \end{abstract}

  
	\end{frontmatter}

\section{Introduction}

Thanks to mobility, miniaturization, and portability, Wearable Mapping System (WMS) has emerged as promising tools in the field of emergency rescue \citep{kachurka2021weco}, safe construction \citep{pirkl2016smart,attard2018tunnel} and Building Information Model (BIM) \citep{yiugit2023investigation,meyer2023geometric}. In particular, wearable 3D mapping systems provide an efficient way to collect 3D information in Global Navigation Satellite System (GNSS)-denied complex environments, which may be difficult to access for common wheeled or legged robots. The 3D information can be used as prior maps for robot navigation \citep{wen2022robust}.

\begin{figure}
\centering
\includegraphics[width=0.8\textwidth]{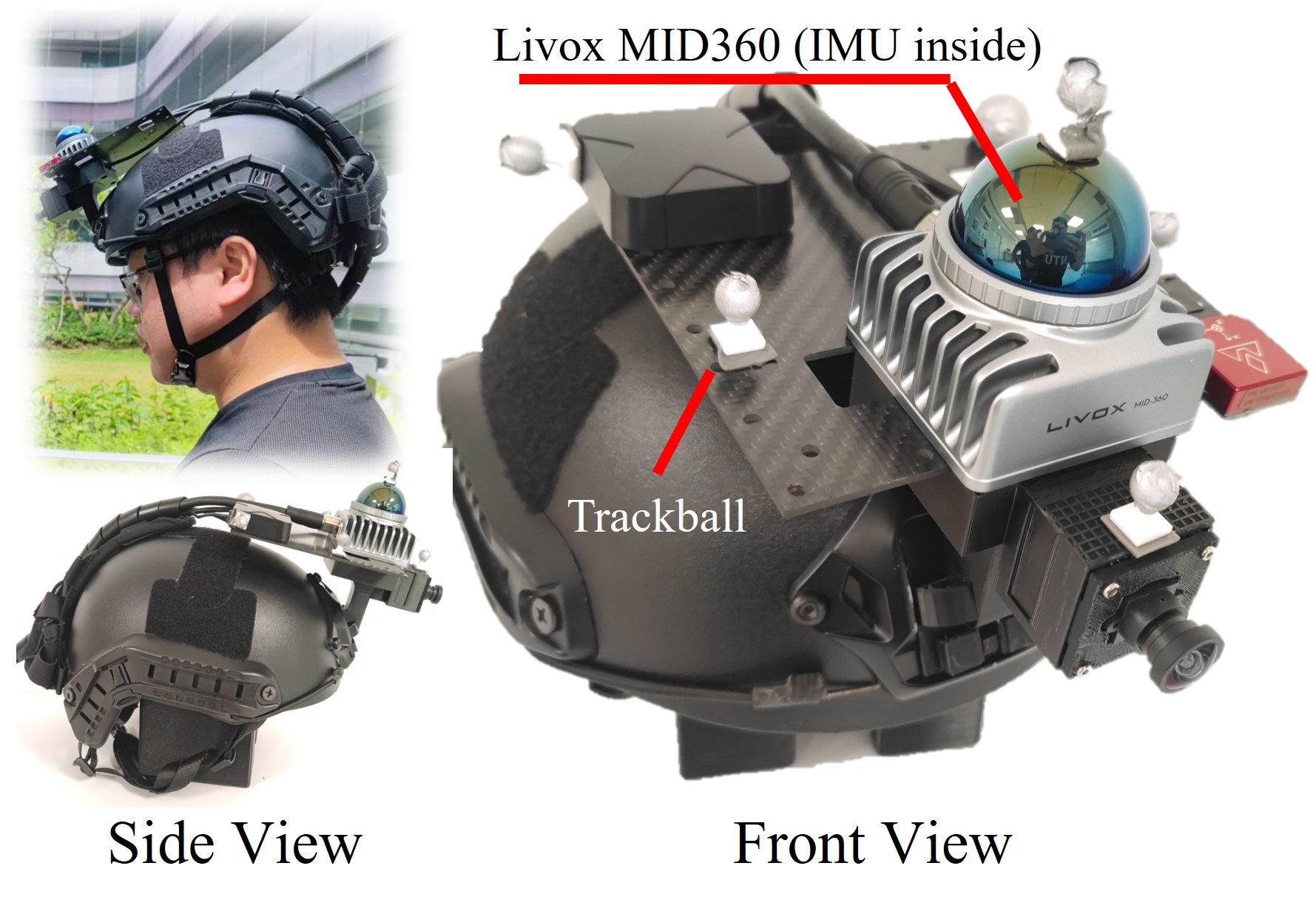}
\caption{Hardware configuration of the compact helmet-based wearable sensing system. The main sensor is the Livox MID360.}
\label{fig_5}
\end{figure}

Several commercial WMSs have been successfully applied in developing prior map for GNSS-denied environments. Some notable examples are the backpack mapping system, NavVis, which combines multiple scanners, and the hand-held mapping system, GeoSLAM, which employs a rotation module. \citet{zhou2023backpack} designed a dual-LiDAR backpack mapping system, which utilizes multiple
ground constraints to restrict the SLAM drift in multi-level buildings. While the multi-scanner configuration and the rotation module can increase the Field of View (FoV), they also increase the weight and burden for the operator. Moreover, existing WMSs usually employ high-grade IMU, which incurs high costs that can be inhibitory for non-professional users. Thus, developing a compact WMS is a hot topic in both academia and industry \citep{li2023whu}. In reducing cost, the limited FoV of single-LiDAR configurations and high noise of low-cost IMUs (e.g., ICM40609 costs only 10 dollars) then pose a challenge for WMS. Moreover, existing WMSs generally require that users move slowly to avoid vibration and improve LiDAR Inertial Odometry (LIO) accuracy \citep{zhang2019slam}, which is inconvenient and also hard to guarantee in the most practical scenarios. Although the effects of vibration are more obvious on a helmet system compared to backpack or handheld ones \citep{beauregard2006helmet}, head-mounted sensing has the benefit of “what you see is what you get” \citep{sidaoui2019collaborative}. In this paper, to boost the capability of compact WMS and satisfy the prior map requirements of automated robot delivery in complex scenes, we are concerned with three aspects of LIO on the compact helmet-based WMS: (1) vibration-tolerant continuous-time optimization based on hybrid IMU factors, (2) human motion-based factors, and (3) feature selection for low-latency and low-drift in complex scenes.

\begin{figure*}
\centering
\includegraphics[width=0.96\textwidth]{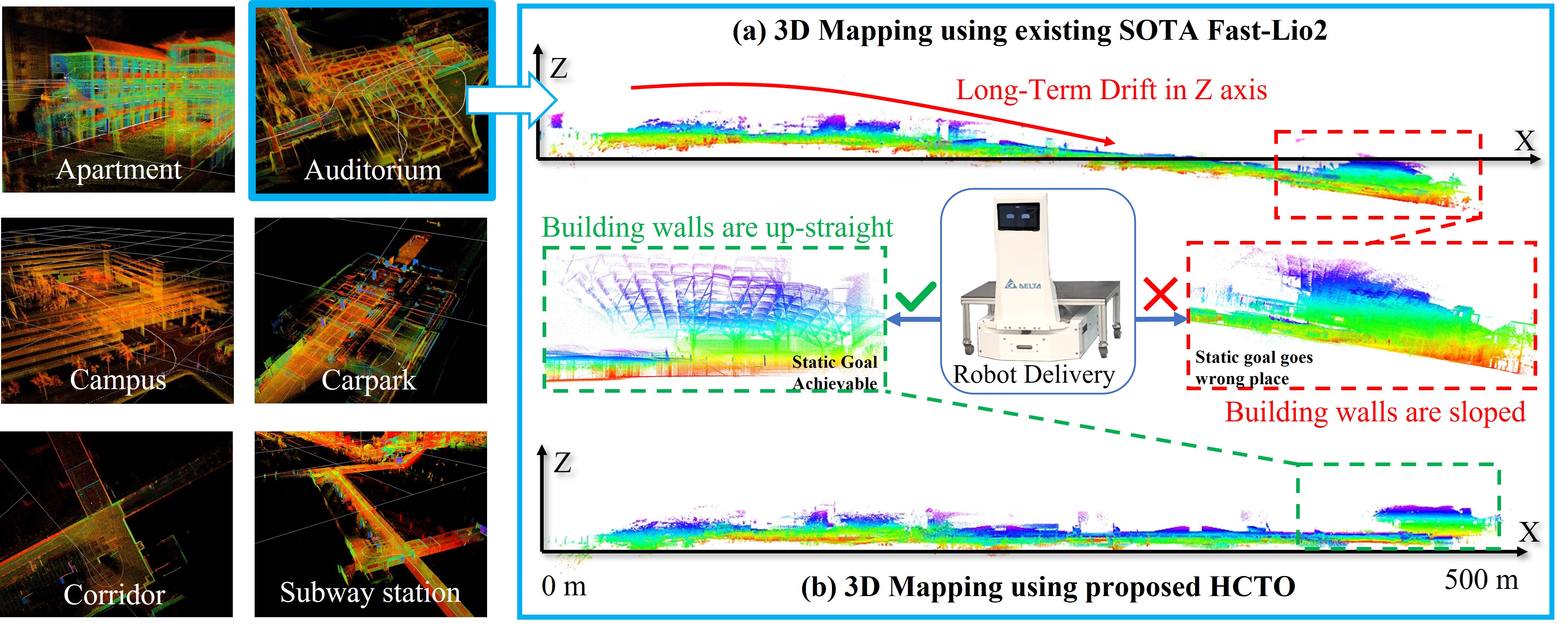}
\caption{ Prior map constructed by Hybrid Continuous Time Optimization (HCTO) for robot navigation in different challenging scenes using wearable devices. (a) The mapping results from Fast-Lio2 \citep{xu2022fast} are corrupted by long-term drift, where the building walls are slanted and the ground plane bends downwards. The robot goes to the wrong place using the map. (b) The mapping results using the proposed HCTO maintain good accuracy, where the building walls are vertical and the ground plane maintains levelness. The robot goes to the right goal.}
\label{fig_1}
\end{figure*}

For trajectory formulation and optimization of LIO, existing methods can be generally categorized into two types: Continuous Time Optimization (CTO) and Discrete Time Optimization (DTO) \citep{cioffi2022continuous}. The traditional DTO has been thoroughly studied \citep{cadena2016past,karam2021simple}, which is not the main target of this paper. Compared with DTO, CTO has several intrinsic benefits and has been investigated for LIO in recent years \citep{nguyen2023slict,lv2021clins}. First, CTO’s trajectory could be sampled at any time, which makes it easy to fuse asynchronous data (LiDAR and IMU), and especially human motion constraints in our WMS case. Second, we do not need to add additional variables at the time instances corresponding to a measurement or motion constraint, which keeps the number of variables manageable. However, CTO assumes polynomial smoothness of the motion, which may lead to loss of information and an ill-fitted model at high-variation segments, i.e., the motion vibration caused by human motions in our WMS case. In this work, to cope with the motion vibration of humans when using CTO, we divide the walking motion into high-frequency parts, low-frequency parts, and constant-velocity parts, then handle them separately. Instead of using a higher-order B-spline trajectory or adding more knots to increase the model fitting ability \citep{cioffi2022continuous}, our method constructs pre-integration constraints and raw measurement constraints in high- and low-frequency parts, respectively. Thus, we refer to the proposed method as hybrid optimization. Moreover, the constant velocity factor is also constructed to limit the attitude drift.

In the aspect of motion-based constraints, zero-velocity detection and update (ZUPT) is the most common human motion constraint \citep{wang2019study}. However, ZUPT is only available for foot-mounted IMU. With the rapid development of deep learning, a lot of data-driven motion-based methods are proposed to predict IMU bias \citep{buchanan2022deep} or velocity \citep{buchanan2022learning} to contain the low-cost IMU error accumulation. Moreover, the motion-based constraints can be integrated into batch optimization to enhance the system performance \citep{clark2017vinet}. The activities detected from the IMU are also used for loop closure successfully \citep{zhou2023asl}. However, in this paper, we do not focus on complex learning of the motion constraint, instead, we provide a continuous fusion framework with a simple yet effective constraint to correct attitude drift. In theory, the attitude (roll and pitch) is observable for LIO and Visual Inertial Odometry (VIO), because of the gravity vector \citep{qin2018vins}. Unfortunately, in practice, due to the uneven distribution or degeneration of feature correspondences \citep{pan2021mulls}, the Iterative Closest Point (ICP)\citep{segal2009generalized}-like the matching process in LIO often converges to a local minimum, further resulting in wrong IMU state estimation and the loss of attitude observability. The loss of attitude observability is common when only a low-cost IMU is used, or the system is moving under vibration according to our experience. In this work, inspired by the complementary filters for orientation estimation \citep{mahony2008nonlinear}, we analyze human constant-velocity motion patterns using the sequential IMU and correct the attitude at each pedestrian step.

For a $k^{th}$ order B-spline trajectory, each constraint corresponds to $k$ knots, which means more Jacobian estimation than the DTO when performing the batch optimization \citep{park2021elasticity}. The maintenance of all feature points in the LIO system may undermine losing the real-time performance of the LIO system. Selecting fewer but more important feature points will boost the performance of the B-spline trajectory estimation. Feature point selection has been widely studied in the field of SLAM \citep{jiao2021greedy,zhao2020good}. LOAM \citep{zhang2014loam} selected features according to space distribution. MULLS \citep{pan2021mulls} selected features considering different semantic types. M-LOAM \citep{jiao2021greedy} selected features according to the information matrix's spectral attributes using a greedy manner. Different from \cite{jiao2021greedy}, we guarantee the optimality of the spline knots (other than pose) considering the similarity between jacobians of LiDAR correspondences. Additionally, we propose a group-based stochastic greedy solver aimed at accelerating the feature selection process through an analysis of feature similarity.

To sum up, applying existing LIO methods on compact WMS mainly faces two challenges: (1) The human motion vibration cannot be modeled using low-order spline (e.g., the commonly used $4^{th}$ order), while a higher-order B-spline or more knots will dramatically increase the computational burden and uncertainty of the system. (2) The uneven or degenerated LiDAR feature correspondences cause the matching process to converge to a local minimum, further resulting in a long-term drift. The above problems result in attitude and Z-direction drift as shown in Fig.\ref{fig_1} (a). The “bent” point cloud map, though geometrically consistent at the local scale, cannot serve as the prior map for the planning and navigation of delivery robots.

Aiming at these problems, we propose a novel LIO with Hybrid Continuous Time Optimization (HCTO) to build an accurate point cloud map in real time. The main contributions of our work are as follows:
\begin{itemize}
    \item HCTO recognizes the Low-Frequency Part (LFP) and High-Frequency Part (HFP) from sequential IMU measurements. IMU constraints based on raw measurements are used in LFP to better fit the observations. On the contrary, IMU pre-integration constraints are used in HFP to suppress severe vibrations, which cannot be fitted using a low-order spline trajectory model. Hence, HCTO realizes efficient fusion of IMU without increasing variables and uncertainty of a spline trajectory.
    \item HCTO recognizes the Constant Velocity Part (CVP) from sequential IMU measurements and corrects system attitude by constructing CVP constraints for the spline trajectory, which restricts the attitude drift.
    \item HCTO proposes a feature selection scheme using optimal design and a group-based stochastic-greedy solver, which achieves real-time performance and better odometry accuracy in degenerated scenes.
    \item HCTO’s validation and effectiveness are demonstrated with extensive experiments on public and in-house helmet-based datasets.
\end{itemize}

The rest of this paper is structured as follows. The preliminaries for our system are provided in Section \ref{section_pre}. A brief system overview including the hardware and software structure is provided in Section \ref{section_sys}. A detailed description of our system is presented in Section \ref{section_method}. The experiments are conducted on public and in-house helmet-based datasets in Section \ref{section_exp}. Conclusion and future work are drawn in Section \ref{section_conclusion}.

\section{Preliminary \label{section_pre}}
\subsection{Notation}

In this paper, for a vector $\mathbf{p}\in\mathbb{R} ^3$, the hat notation $\hat{\mathbf{p}}$ denotes the global optimization-based estimation of $\mathbf{p}$. The breve notation $\Breve{\mathbf{p}}$ denotes the IMU-propagated information. For a matrix $\mathbf{M}$, $\mathbf{M}_{r,c}$ denotes the element at $r^{th}$ row and  $c^{th}$ column of $\mathbf{M}$. The orientation could be represented by rotation matrix $\mathbf{R}\in SO(3)$. We reserve the right superscript and subscript for the coordinate transformation, i.e.,  $\left[ \mathbf{R}_{B}^{A},\mathbf{p}_{B}^{A} \right]$ transform a vector from frame B to frame A.
\subsection{State Estimation}

We define the $i^{th}$ time segment spanning a period $[t_i,t_{i+1})$ with the constant duration $\Delta t$. The system state $\mathcalbf{X}(t)$ at time $t\in[t_i,t_{i+1}) $ is defined as follows:

\begin{equation}
\mathcalbf{X} \left( t \right) =\left[ \mathbf{R}_{b}^{w}\left( t \right) ,\mathbf{p}_{b}^{w}\left( t \right) ,\mathbf{b}_a\left( t \right) ,\mathbf{b}_g\left( t \right) \right] \in SO\left( 3 \right) \times \mathbb{R} ^9, \label{eq1}
\end{equation}

where $\mathbf{R}_{b}^{w}(t),\mathbf{p}_{b}^{w}\left( t \right)$ are rotation and translation of body frame defined in the world frame. $\mathbf{q}_{b}^{w}(t)$ is the quaternion corresponding to $\mathbf{R}_{b}^{w}(t)$.  $\mathbf{b}_a\left( t \right),\mathbf{b}_g\left( t \right)$ are IMU accelerometer and gyroscope biases, which are modeled using the random-walk model for low-cost IMU \citep{shin2004unscented}. It should be noted that the velocity $\mathbf{v}_{b}^{w}\left( t \right)$ is not included in the system states as we use the B-spline to formulate the trajectory.

\subsection{IMU pre-integration}
IMU pre-integration is commonly used for DTO by integrating the IMU measurements in time duration to get relative spatial relationships. Given the $i^{th}$ time segment $t\in [t_i,t_{i+1} )$, the state variables are constrained by the IMU measurements (acceleration $\breve{\mathbf{a}}^b(t)$ and gyroscope $\breve{\mathbf{\omega}}^b(t)$) in this time segment with IMU integration as follows:
\begin{equation}
\begin{aligned}
\mathbf{p}_{b}^{w}\left( t_{i+1} \right) &=\mathbf{p}_{b}^{w}\left( t_i \right) +\mathbf{v}_{b}^{w}\left( t_i \right) \Delta t
\\
&+\iint_{t\in \left[ t_i,t_{i+1} \right)}{\left( \mathbf{R}_{b_t}^{w}\left( \breve{\mathbf{a}}^b(t)-\mathbf{b}_a\left( t_i \right) \right) -\mathbf{g}^w \right)}dt^2,
\\
\mathbf{v}_{b}^{w}\left( t_{i+1} \right) &=\mathbf{v}_{b}^{w}\left( t_i \right)
\\
&+\int_{t\in \left[ t_i,t_{i+1} \right)}{\left( \mathbf{R}_{b_t}^{w}\left( \breve{\mathbf{a}}^b(t)-\mathbf{b}_a\left( t_i \right) \right) -\mathbf{g}^w \right) dt},
\\
\mathbf{q}_{b}^{w}\left( t_{i+1} \right) &=\mathbf{q}_{b}^{w}\left( t_i \right) 
\\
&\otimes \int_{t\in \left[ t_i,t_{i+1} \right)}{0.5\mathbf{\Omega }\left( \breve{\mathbf{\omega}}^b(t)-\mathbf{b}_g\left( t_i \right) \right) \mathbf{q}_{b_t}^{w}dt},
\end{aligned}
\end{equation}
where
\begin{equation}
\begin{aligned}
\mathbf{\Omega }\left( \mathbf{\omega } \right) =\left[ \begin{matrix}
	-\mathbf{\omega }_{\times}&		\mathbf{\omega }\\
	-\mathbf{\omega }^{\top}&		0\\
\end{matrix} \right] ,\mathbf{\omega }_{\times}=\left[ \begin{matrix}
	0&		-\mathbf{\omega }_2&		\mathbf{\omega }_1\\
	\mathbf{\omega }_2&		0&		-\mathbf{\omega }_0\\
	-\mathbf{\omega }_1&		\mathbf{\omega }_0&		0\\
\end{matrix} \right], 
\end{aligned}
\end{equation}

where $\mathbf{g}^w$ is the gravity vector in the world frame. $\mathbf{\omega }_0$, $\mathbf{\omega }_1$, and $\mathbf{\omega }_2$ are the three elements in vector $\mathbf{\omega}$. After changing the reference frame of integration to $b_{i}$, the three IMU pre-integration factors, $\Delta \mathbf{p}_{i+1}^{i}, \Delta \mathbf{v}_{i+1}^{i}$ and $\Delta \mathbf{q}_{i+1}^{i}$, are defined in Eq.\eqref{eq_preinte_sections} \citep{qin2018vins}. These three factors can also be used to restrict the relative motion of the spline trajectory, which will be introduced in the following subsections.
\begin{equation}
\begin{aligned}
\Delta \mathbf{p}_{i+1}^{i}&=\iint_{t\in \left[ t_i,t_{i+1} \right)}{\mathbf{R}_{b_t}^{b_i}\left( \breve{\mathbf{a}}_t-\mathbf{b}_a\left( t_i \right) \right)}dt^2,
\\
\Delta \mathbf{v}_{i+1}^{i}&=\int_{t\in \left[ t_i,t_{i+1} \right)}{\mathbf{R}_{b_t}^{b_i}\left( \breve{\mathbf{a}}_t-\mathbf{b}_a\left( t_i \right) \right) dt},
\\
\Delta \mathbf{q}_{i+1}^{i}&=\int_{t\in \left[ t_i,t_{i+1} \right)}{0.5\mathbf{\Omega }\left( \breve{\mathbf{\omega}}_t-\mathbf{b}_g\left( t_i \right) \right) \Delta \mathbf{q}_{b_t}^{b_i}dt}.\label{eq_preinte_sections}
\end{aligned}
\end{equation}

\subsection{Continuous Time Optimization (CTO) using B-Spline}
A $k^{th}$ order B-Spline with $N+1$ knots (or control points)   could be formulated using the De Boor-Cox recurrence relation as follows:
\begin{equation}
\mathbf{p}\left( t \right) =\sum_{i=0}^N{B_{i,k}(t)\mathbf{p}_i},\label{eq2}
\end{equation}
where $B_{i,k}(t)$ is the B-Spline basis function \citep{qin1998general}. We define $u(t) :=( t-t_0) /\Delta t-i$ as normalized time elapsed in the window $[t_i,t_{i+1})$ and write it as $u$ for simplification. The value 
$\textbf{p}(u)$ could be represented by matrix multiplication:
\begin{equation}
\mathbf{p}\left( u \right) =\left[ \mathbf{p}_i,\mathbf{p}_{i+1},\cdots ,\mathbf{p}_{i+k-1} \right] \mathbf{Mu},
\label{eq3}
\end{equation}
where $\mathbf{u}_n=u^n$. $\textbf{M}$ is the $k^{th}$ order blending matrix \citep{qin1998general}:
\begin{equation}
\begin{split}
\mathbf{M}_{s,n}=\frac{C_{k-1}^{n}}{\left( k-1 \right) !}\sum_{l=s}^{k-1}{\left( -1 \right) ^{l-s}\left( k-l-1 \right) ^{k-1-n}},\\
s,n \in { 0,...,k-1 }.\label{eq4}
\end{split}
\end{equation}
$\textbf{p}(u)$ can also be transformed to a cumulative formulation \citep{sommer2020efficient}:

\begin{equation}
\begin{split}
\mathbf{p}\left( u \right) =\mathbf{p}_i+\sum_{j=1}^{k-1}{\left( \mathbf{p}_{i+j}-\mathbf{p}_{i+j-1} \right) (\widetilde{\mathbf{M}}\mathbf{u}})_j,\label{eq5}
\end{split}
\end{equation}
where $\widetilde{\mathbf{M}}_{j,n}=\sum\nolimits_{s=j}^{k-1}{\mathbf{M}_{s,n}}$. As our system state knot $\mathcalbf{X} _i$
is in a Lie group, the cumulative B-Spline of the system with state knots $\left\{ \mathcalbf{X} _0,...\mathcalbf{X} _i,...,\mathcalbf{X} _N \right\}$ has the following form:
\begin{equation}
\begin{split}
\mathcalbf{X} \left( u \right) =\mathcalbf{X} _i\prod_{j=1}^{k-1}{\mathrm{Exp}\left( \mathrm{Log}\left( \mathcalbf{X} _{i+j-1}^{-1}\mathcalbf{X} _{i+j} \right) (\widetilde{\mathbf{M}}\mathbf{u})_j \right)}.\label{eq6}
\end{split}
\end{equation}

We can also obtain the velocity $\dot{\mathbf{p}}_{b}^{w}(u)$
 , acceleration $\ddot{\mathbf{p}}_{b}^{w}(u)$, and angular velocity $\dot{\mathbf{R}}_{b}^{w}\left( u \right)$ at any time according to the differential of $\mathcalbf{X}(u)$. Using the IMU measurements at these specific times, we could restrict the spline trajectory by comparing the differential results and raw IMU measurements, which is the standard way to fuse the IMU information of the CTO \citep{cioffi2022continuous}.

\subsection{Maximum-A-Priori (MAP) Optimization in a sliding window}

\begin{figure}[h]
    \centering
    \includegraphics[width=0.8\textwidth]{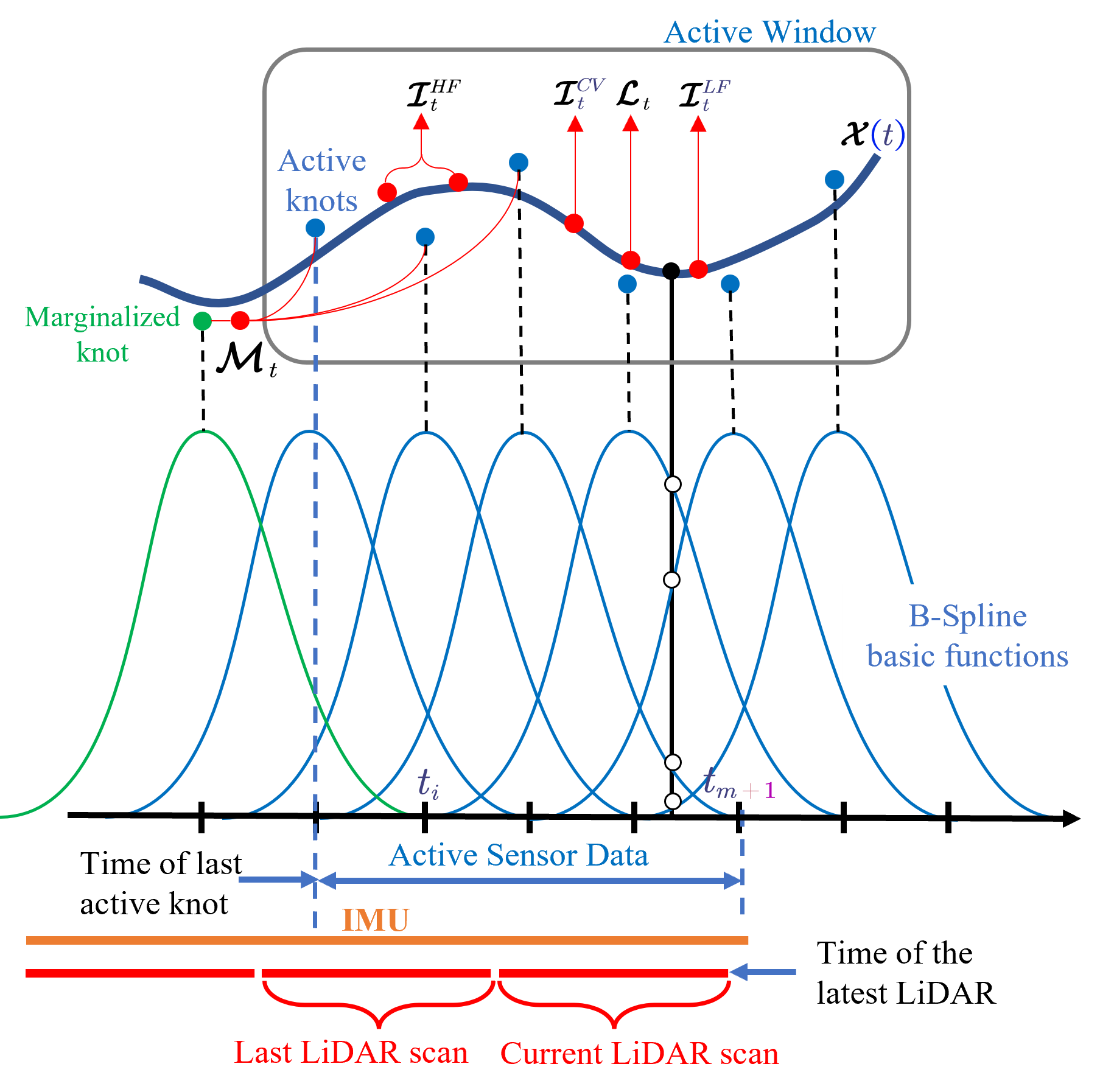}
    \caption{Maximum-A-Priori (MAP) optimization in the active window taking the $4^{th}$ order B-Spline as an example.}
    \label{fig_4}
\end{figure}

The core of the proposed LIO is solving the following MAP problem as illustrated in  Fig. \ref{fig_4} and written as Eq. \ref{eq_map}:

\begin{equation}
\begin{aligned}
f(\hat{\mathcalbf{X}}) =\sum_{t\in \mathbf{\Phi }}{\left\| r_{t}^{\mathcal{H} \mathcal{F}} \right\|_{\Sigma_{\mathcal{HF}}}^2}+\sum_{t\in \mathbf{\Phi }}{\left\| r_{t}^{\mathcal{L} \mathcal{F}} \right\|_{\Sigma_{\LF}}^2}
\\
+\sum_{t\in \mathbf{\Phi }}{\left\| r_{t}^{\mathcal{C} \mathcal{V}} \right\|^2_{\Sigma_{\mathcal{CV}}}}+\sum_{t\in \mathbf{\Phi }}{\left\| r_{t}^{\mathcal{L}} \right\|^2_{\Sigma_{\mathcal{L}}}}+\left\| r_{t}^{\mathcal{M}} \right\|^2_{\Sigma_{\mathcal{M}}}.
\label{eq_map}
\end{aligned}
\end{equation}

Our system solves the above $k^{th}$ order B-Spline MAP problem every time receiving a LiDAR frame. Assume $t_{i}$ and $t_{i+m}$, just cover the lower bound and upper bound for the current LiDAR scan. According to the locality of B-Spline, the trajectory segment of $(t_{i}, t_{i+m})$ is affected by a set of knots $[\mathcalbf{X}_{i}, ..., \mathcalbf{X}_{i+m+k-1}]$, defined as active nodes \citep{lv2021clins}. Only the active knots will be optimized in the current sliding window. We define the active sensor data duration as $\mathbf{\Phi}=[t_i, t_{i+m}]$, and use the factors $r_{t}^{\mathcal{CV}}$ (CVP factor defined in Section \ref{4.2.1}), $r_{t}^{\mathcal{HF}}$ (HFP factor defined in Section \ref{4.2.2}), $r_{t}^{\LF}$ (LFP factor defined in Section \ref{4.2.3}), and $r_{t}^{\mathcal{L}}$ (LiDAR factor defined in Section \ref{4.3}) within the duration $\mathbf{\Phi}$ for the optimization. We also marginalize the knot out of the window to build a prior constraint $r_{t}^{\mathcal{M}}$ for the first $k-1$ knots in the active window to achieve better accuracy while restricting the computation load \citep{lv2023continuous}. $\Sigma_{\mathcal{CV}},\Sigma_{\mathcal{HF}},\Sigma_{\LF},\Sigma_{\mathcal{L}}$, and $\Sigma_{\mathcal{M}}$ are the covariance matrices corresponding to the factors. $\Sigma_{\mathcal{CV}},\Sigma_{\mathcal{HF}},\Sigma_{\mathcal{LF}}$ are related to the raw measurements of IMU, which could be set according to the measurement noises from the IMU specifications. $\Sigma_{\mathcal{L}}$ is set according to the LiDAR measurement accuracy (3 cm for most Livox LiDAR). $\Sigma_{\mathcal{M}}$ is calculated according to the marginalization \citep{qin2018vins}. It should be noticed that, like most LIO systems \citep{xu2022fast}, we propagated the IMU measurement to get the initial poses. While the system states presented in knots of B-Spline is initialized with IMU-only optimization \citep{lv2021clins}.

\section{System Overview \label{section_sys}}

\subsection{Hardware of the compact helmet-based sensing system}

We integrated a compact helmet-based sensing system as illustrated in Fig.\ref{fig_5}. The main sensor is the Livox MID360, which contains a built-in low-cost IMU (ICM40609). The sensors are mounted on a standard Future Assault Shell Technology (FAST) Helmet. The weight of the wearable sensing system is about 1.2 kg.

\subsection{Main function blocks of diagram of the proposed helmet-based LIO}

\begin{figure*}
\centering
\includegraphics[width=5.5in]{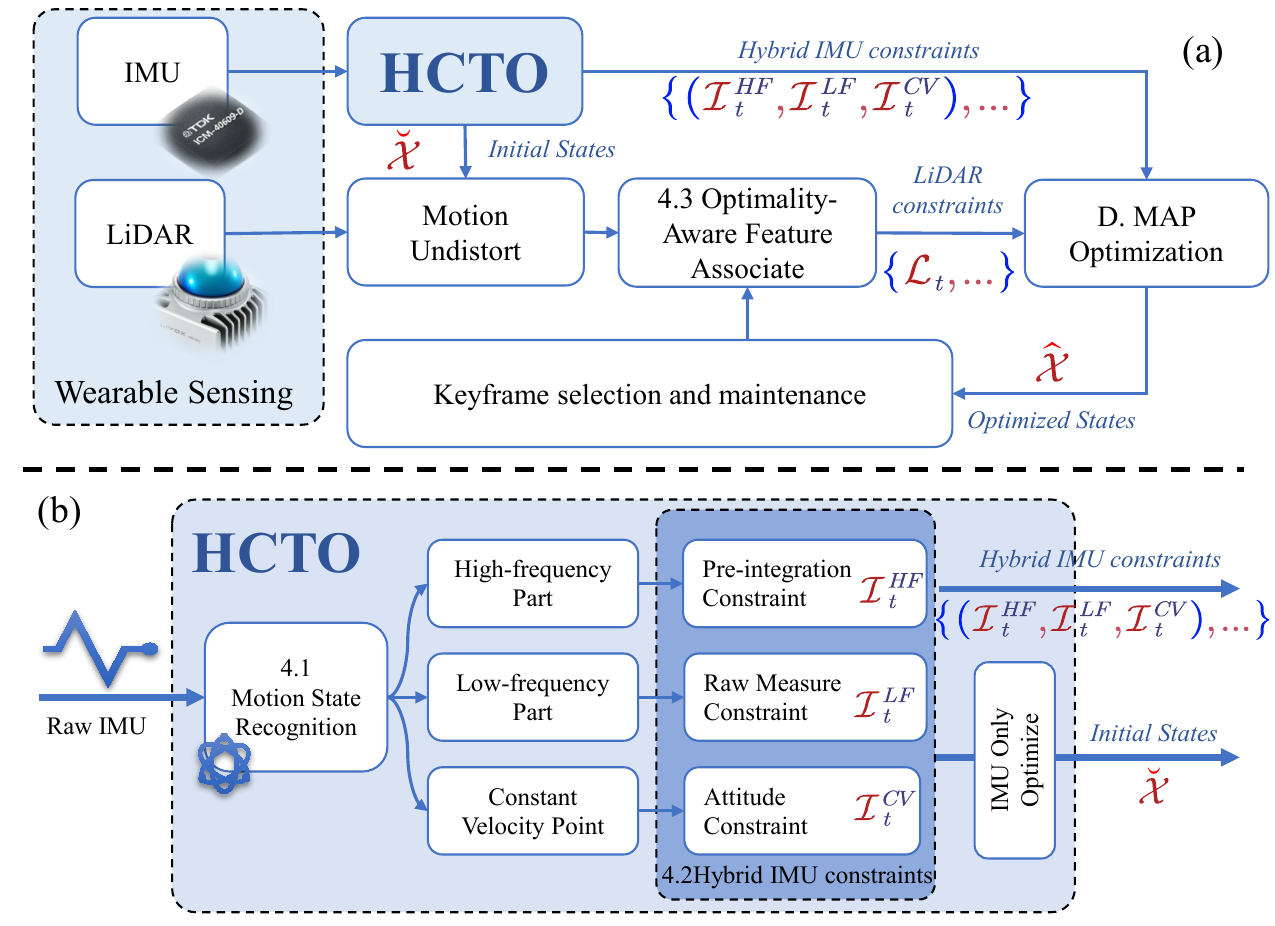}
\caption{System overview of the proposed helmet-based LIO system. (a) LiDAR Inertial Odometry (LIO) with Hybrid Continuous Time Optimization (HCTO). (b) Diagram of HCTO.}
\label{fig_2}
\end{figure*}

Fig. \ref{fig_2} provides an overview of our helmet-based LIO system. Taking the low-cost IMU and solid-state LiDAR data as input, the proposed system constructs hybrid IMU constraints while removing the LiDAR motion distortion based on HCTO (Section \ref{4.1} and \ref{4.2}) first. Then, the point-to-plane correspondences are extracted and selected using the optimal-aware point feature association block (Section \ref{4.3}). The system states are optimized using hybrid IMU factors and LiDAR factors in a sliding window when a new LiDAR frame comes. At last, the current LiDAR frame is maintained in the keyframe database if it is selected as a keyframe (Section \ref{4.4}). In the next section, we detail the numbered blocks. 

\section{Methodology \label{section_method}}

\subsection{Motion state recognition\label{4.1}}

\begin{figure}[h]
    \centering
    \includegraphics[width=0.8\textwidth]{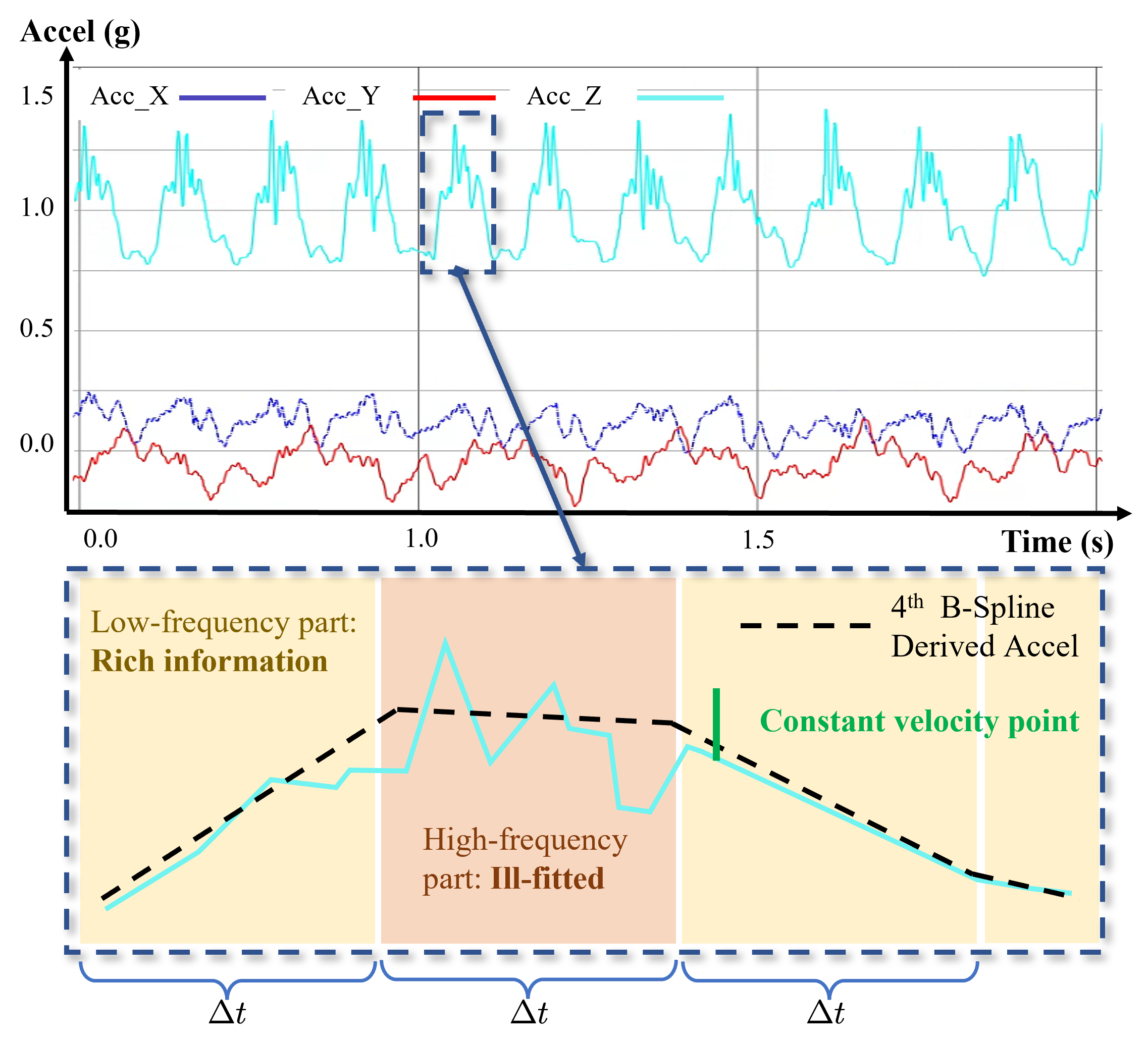}
    \caption{Illustration of different motion states in the repetitive motion pattern.}
    \label{fig_3}
\end{figure}

Fig. \ref{fig_3} illustrates the accelerometer output on the helmet during human motion. It can be easily observed that there is a repetitive motion pattern for each step. As for the most commonly used cubic Spline ($4^{th}$ order), it only guarantees the linear changes of acceleration, which results in ill-fitting using the raw IMU measurements as residual during the vibration, as shown in the zoomed view of Fig. \ref{fig_3}. 
The LFP segment can be effectively modeled using a linear model, whereas the HFP segment cannot. Accordingly, we partition the trajectory into two parts, HFP and LFP, delineated by the Root Mean Squared Error (RMSE) $\sigma_{motion}$ of acceleration measurements within the B-Spline time interval $\Delta t$, as derived from linear fitting residuals.
If $\sigma_{motion}$ is bigger than three times the accelerometer measurement error, this time segment is determined as an HFP following the three-sigma rule. Otherwise, this time segment belongs to LPF because the acceleration measurements can be fitted well using a linear model.

Moreover, we can obtain the initial guess of acceleration $ \mathbf{a}^w\left( t \right)$ in the world frame using pure IMU integration at any time $t$. We detect the time $\left\| \mathbf{a}^w\left( t \right) \right\| = 0$, which is the moment called the CVP. Details on how to use them to build hybrid IMU factors are in the following section.

\subsection{Construction of Hybrid IMU factors \label{4.2}}
\subsubsection{CVP\label{4.2.1}} 
Inspired by the complementary filters for orientation estimation \citep{mahony2008nonlinear}, we first investigate the force equation of the system as follows:
\begin{equation}
\begin{split}
\mathbf{R}_{b}^{w}\left( t \right) \left( \mathbf{a}^b(t) +\mathbf{b}_a\left( t \right) \right) +\mathbf{g}^w=\mathbf{a}^w\left( t \right).\label{eq7}
\end{split}
\end{equation}
As $\left\| \mathbf{a}^w\left( t \right) \right\| =0$, we have the following relationship between the states and the gravity:
\begin{equation}
\mathbf{R}_{b}^{w}\left( t \right) \left( \mathbf{a}^b(t) +\mathbf{b}_a\left( t \right) \right) =-\mathbf{g}^w,\ 
\text{if } \left\| \mathbf{a}^w\left( t \right) \right\| = 0. \label{eq: const velocity model}
\end{equation}
We use the condition $\norm{\breve{\mathbf{a}}^w(t)} < 0.3$ as the trigger for adding a CVP factor to Eq. \eqref{eq7}, where $\breve{\mathbf{a}}^w(t)$ is the IMU-predicted acceleration in the world frame Eq. \eqref{eq: const velocity model}.

We only use the CVP factor to correct the roll and pitch of the state estimate.
Assume $\delta \mathbf{\theta }(t)$ is the rotation correction corresponding to initial guess  $\breve{\mathbf{R}}_{b}^{w}(t)$, Eq. \eqref{eq: const velocity model} can be written as:
\begin{equation}
\begin{split}
-\breve{\mathbf{R}}_{b}^{w}\left( t \right) ^{\top}\left( I-\left[ \delta \hat{\mathbf{\theta}}\left( t \right) \right] _{\times} \right) \mathbf{g}^w =  \breve{\mathbf{a}}^b(t) +\mathbf{b}_a(t).
\end{split}
\end{equation}
Then the constant velocity factor $r_{t}^{\mathcal{CV}}$ at time t is written as follow:
\begin{equation}
\begin{split}
r_{t}^{\mathcal{CV}} &=  
\left[ \begin{matrix}
	0&		-1\\
	1&		0\\
\end{matrix} \right] \left[ \begin{array}{c}
	\delta \mathbf{\hat \theta }\left( t \right) _0\\
	\delta \mathbf{\hat \theta }\left( t \right) _1\\
\end{array} \right] 
\\
+& \frac{1}{\left\| \mathbf{g}^w \right\|}\left[ \left( \breve{\mathbf{R}}_{b}^{w}\left( t \right) ^{\top} \breve{\mathbf{a}}^b(t) +\mathbf{b}_a\left( t \right) \right) +\breve{\mathbf{R}}_{b}^{w}\left( t \right) ^{\top}\mathbf{g}^w \right] _{1:2}.
\end{split}
\end{equation}
where $\delta \mathbf{\hat \theta }(t)_0$ and $\delta \mathbf{\hat \theta }(t)_1$ are the roll and pitch corrections, respectively. 

\subsubsection{HFP\label{4.2.2}}

As illustrated in Fig.\ref{fig_3}, the high-frequency part of motion can not be fitted using $4^{th}$ order B-Spline, which only guarantees the linear changing of acceleration, hence cannot represent the high-frequency motion. Consequently, naively fitting this will lead to a large error and unstable results. Increasing the order of B-Spline or decreasing the $\Delta t$, namely increasing the parameter of the trajectory, could be one solution but will increase the computation load and uncertainty of the system. Instead of increasing the parameter to fit the vibrated motion, we integrate the raw IMU measurements and construct HPF factors to constrain the relative motion between the start and end of the high-frequency motion, which alleviates the effect of the high-frequency part on sequential B-Spline segments.
Assume the period $[t_i,t_{i+1})$ with duration $\Delta t$ belongs to HFP. We integrate the IMU measurements in $[t_i,t_{i+1})$ to get the position, velocity, and rotation elements ($\Delta \mathbf{p}_{i+1}^{i}, \Delta \mathbf{v}_{i+1}^{i}$ and $ \Delta \mathbf{q}_{i+1}^{i}$) of IMU pre-integration according to Eq. \eqref{eq_preinte_sections}. The HFP factor $r_{t}^{\mathcal{HF}}$ is written as follows:
\begin{equation}
\begin{split}
&r_{t}^{\mathcal{HF}}=
\\
&\left[ \begin{array}{c}
	\Delta \mathbf{p}_{i+1}^{i}\\
	\Delta \mathbf{v}_{i+1}^{i}\\
	\Delta \mathbf{q}_{i+1}^{i}\\
\end{array} \right] - 
\left[ \begin{array}{c}
	\mathbf{R}_{b}^{w}\left( t_i \right) ^{\top}\left( \begin{array}{c}
	\mathbf{p}_{b}^{w}\left( t_{i+1} \right) -\mathbf{p}_{b}^{w}\left( t_i \right)\\
	-0.5\mathbf{g}^w\Delta t^2-\dot{\mathbf{p}}_{b}^{w}\left( t_i \right) \Delta t\\
\end{array} \right)\\
	\mathbf{R}_{b}^{w}\left( t_i \right) ^{\top}\left( \dot{\mathbf{p}}_{b}^{w}\left( t_{i+1} \right) -\mathbf{g}^w\Delta t-\dot{\mathbf{p}}_{b}^{w}\left( t_i \right) \right)\\
	\mathbf{R}_{b}^{w}\left( t_i \right) ^{\top}\mathbf{R}_{b}^{w}\left( t_{i+1} \right)\\
\end{array} \right]. 
\end{split}
\end{equation}

\textbf{Remark. 1} It's important to highlight that the system velocity is extracted from the differential of the B-Spline trajectory, setting it apart from the conventional DTO and ensuring local trajectory consistency. In the high-frequency part, unlike the traditional CTO, the B-Spline trajectory doesn't attempt to fit the oscillated acceleration measurements, thereby facilitating stable and swift convergence.

\textbf{Remark. 2} Integration of the raw IMU measurements in the high-frequency part can be regarded as a special low-pass filter \citep{tavares1966comparison}. This integration effectively suppresses the peak values attributed to jitter. However, the HPF factor introduced here differs from mere low-pass filtering applied to the acceleration measurement in one axis. Instead, it incorporates the IMU measurements according to the strap-down IMU integration principle, ensuring the relative motion relationship remains consistent from the beginning to the end of the high-frequency motion.

\subsubsection{LFP\label{4.2.3}}

For the low-frequency portion of the motion, all IMU measurements are directly employed in the B-Spline parameter estimation to fully utilize the available information.
If the period $[ t_i,t_{i+1})$  belongs to LFP, we construct the following LFP factor $r_{t}^{\LF}$ for each raw IMU measurement:
\begin{equation}
\begin{split}
r_{t}^{\LF}=\left[ \begin{array}{c}
	\ddot{\mathbf{p}}_{b}^{w}\left( t \right)\\
	\dot{\mathbf{R}}_{b}^{w}\left( t \right)\\
\end{array} \right] -\left[ \begin{array}{c}
	\mathbf{R}_{b}^{w}\left( t \right) \left( \breve{\mathbf{a}}^b(t) +\mathbf{b}_a\left( t \right) \right)+\mathbf{g}^w\\
	\,\,\breve{\mathbf{\omega}}^b(t) +\mathbf{b}_g\left( t \right)\\
\end{array} \right]. \label{eq12}
\end{split}
\end{equation}

Using only $r_{t}^{\mathcal{CV}}$, $r_{t}^{\mathcal{HF}}$, and $r_{t}^{\LF}$, within a LiDAR frame duration (i.e., 0.1 s in most scanning configuration), we conduct an IMU only optimization to get the initial guess of system states $\breve{\mathcalbf{X}}\left( t \right)$, which is used to correct the LiDAR motion distortion. The correction is important for the following LiDAR constraints association.

\subsection{Optimality-aware LiDAR factor association \label{4.3}}

\begin{figure*}[h]
\centering
\includegraphics[width=\textwidth]{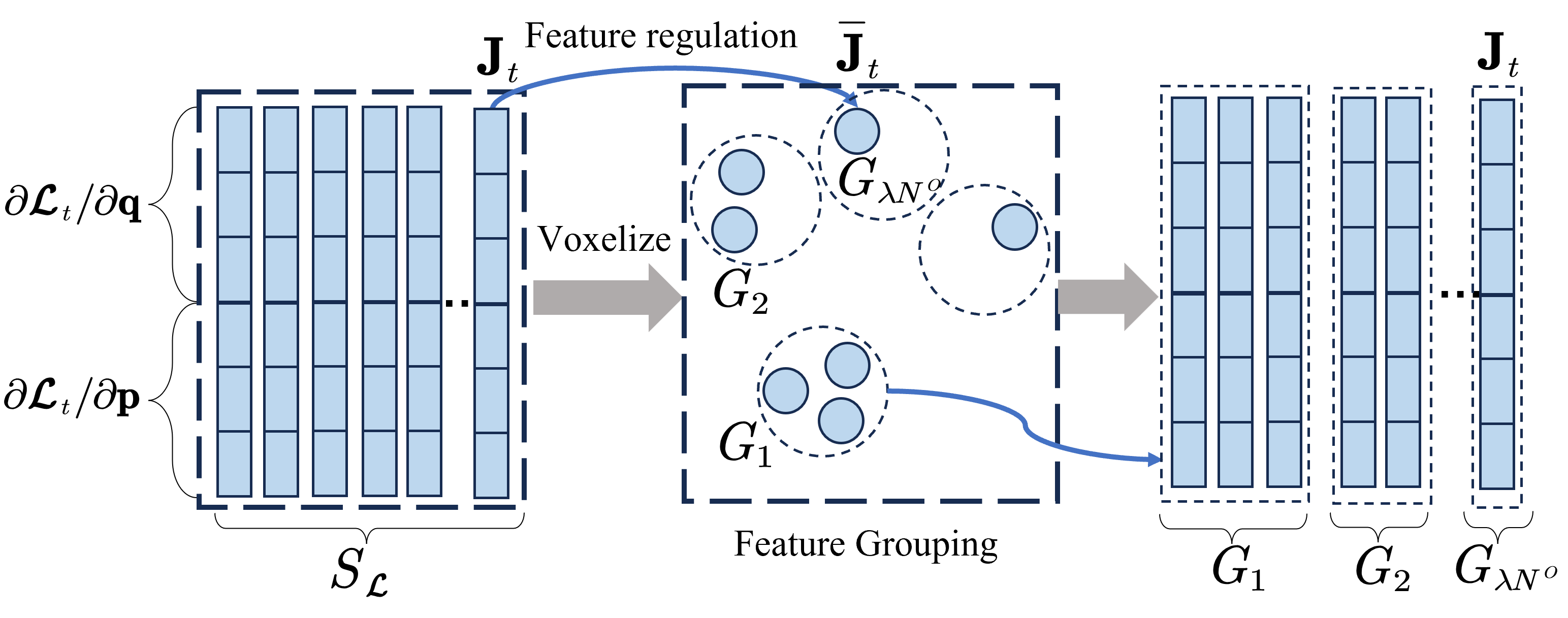}
\caption{Grouping the candidate correspondences in the current LiDAR frame.}
\label{fig_sampling}
\end{figure*}

According to the initial guess of system states $\breve{\mathcalbf{X}}$, we project the points in the current frame into the map coordinate. For a LiDAR measurement $\mathbf{f}_t$, we search its closest points from the map points and calculate their average value $\mu$ and plane normal $\mathbf{n}$. The map points are maintained using the ikdtree \citep{xu2022fast} and will be introduced in Section \ref{4.4}. The point-to-plane factor $r_{t}^{\mathcal{L}}$ is written as follows:

\begin{equation}
\begin{split}
r_{t}^{\mathcal{L}} = \mathbf{n}^{\top}(\mathbf{R}^w_b(t)\mathbf{f}_t+\mathbf{p}^w_b(t)-\mathcalbf{\mu}).
\label{eq13}
\end{split}
\end{equation}

Given the set of point-to-plane factors in the current LiDAR frame $ S_{\mathcalbf{L}} = \{r_{t}^{\mathcal{L}} \}$, there are usually more than several thousand LiDAR correspondences to be added to the MAP optimization in the sliding window, which leads to heavy computation load for B-Spline estimation and unforeseen degeneration caused by the uneven distribution of the correspondences \citep{zhao2020good,zhang2014loam}. Selecting good correspondences for the state estimation not only benefits the time efficiency but also odometry accuracy \citep{jiao2021greedy}. The selection of an optimal correspondence subset $S^{O}_{\mathcalbf{L}} \subset S_{\mathcalbf{L}}$ for variable solving is also called an optimal design \citep{pukelsheim2006optimal}.

As our system states $\mathcalbf{X}$ are presented as $k^{th}$ order B-Spline, hence there are $k$ knots related to the constant $\mathcalbf{L}_t$. As the center knot has the most correlation with the current LiDAR frame according to the B-Spline definition, we only consider the center knot located in the current LiDAR frame time duration for the following optimal analysis to guarantee real-time performance in practice. We linearize $\mathcalbf{L}_t$ and calculate its Jacobian w.r.t $\mathcalbf{X}$ as $\textbf{J}_t= \partial \mathcalbf{L}_t/ \partial \mathcalbf{X}|_{\mathcalbf{X}=\breve{\mathcalbf{X}}} $. The information matrix $\varLambda$ for $S^{O}_{\mathcalbf{L}}$ is defined as follows:

\begin{equation}
\begin{split}
\varLambda = \sum_{L_t\in S^{O}_{\mathcalbf{L}}}{\textbf{J}_{t}^{\top}\textbf{J}_t}.
\end{split}
\end{equation}

The information matrix $\varLambda$ reveals the uncertainties for the variables to be estimated and also the optimality of the subset  $S^{O}_{\mathcalbf{L}}$ \citep{pukelsheim2006optimal}. Specifically, we utilize the minimum eigenvalue $\lambda_{min} (\varLambda)$, a commonly used metric, to qualify the optimality of information matrix $\varLambda$ \citep{jiao2021greedy}. We select the minimum eigenvalue as the metric because it restricts the possible degenerated direction. To achieve the most optimal design for the knots estimation, we try to maximize the objective function $g(S^{O}_{\mathcalbf{L}})$:

\begin{equation}
\begin{split}
arg\max\ &g(S^{O}_{\mathcalbf{L}}) \ s.t.\  S^{O}_{\mathcalbf{L}}\subset S_{\mathcalbf{L}}, |S^{O}_{\mathcalbf{L}}| \leq N^{O}, 
\\
&g(S)=\lambda_{min} \left[ \varLambda \left(S \right) \right],
\label{eq_argmax}
\end{split}
\end{equation}
where $N^{O}$ is the desirable size of correspondences remaining in the system considering the computational load. The objective function in Eq. \eqref{eq_argmax} is an NP-hard problem and could be approximately solved by stochastic-greedy heuristic \citep{khanna2017scalable}. The stochastic-greedy algorithm starts with an empty set. Then, at each step, it selects one element from a random set with a size of $|S_{\mathcalbf{L}}|/N^{O}\mathrm{log}(1/\epsilon) $ from the remaining correspondence set, which gains most of the objective function $g(S^{O}_{\mathcalbf{L}})$. $\epsilon$ is the decay factor. The iterative process stops until it gets $N^{O}$ correspondences. 

\textbf{Theorem. 1} \citep{mirzasoleiman2015lazier} The set function $g(S)$ is submodular and monotone increasing w.r.t $S$.

\textbf{Theorem. 2} \citep{nemhauser1978analysis} Let $S^{O*}_{\mathcalbf{L}}$ and $S^{O\#}_{\mathcalbf{L}}$ be the optimal set and stochastic-greedy heuristic result, respectively, then 

\begin{equation}
\begin{split}
\frac{g(S^{O*}_{\mathcalbf{L}})-g(S^{O\#}_{\mathcalbf{L}})}{g(S^{O*}_{\mathcalbf{L}})} \leq \left(\frac{N^{O}-1}{N^{O}}\right)^{N^{O}}+\epsilon \leq \frac{1}{e} + \epsilon.\label{eq_worst_bound}
\end{split}
\end{equation}

\begin{algorithm}
\DontPrintSemicolon
  \SetAlgoLined
  \KwIn { $S_{\mathcalbf{L}}$, ${\{\mathbf{J}_t\}}$, $\epsilon$, $\lambda$, $N^{O}, D_{v}$}
  \KwOut { Selected LiDAR correspondences $S^{O}_{\mathcalbf{L}}$}
  Initialization of the set  $S^{O}_{\mathcalbf{L}} \gets \oslash $;\;
  Calculate mean value $\{\mu_0,\mu_1,...,\mu_5\}$ and variance  $\{\Sigma_0,\Sigma_1,...,\Sigma_5\}$ at each dimension for ${\{\mathbf{J}_t\}}$;\;
 
  \ForEach {$\mathbf{J}_t \in \{ \mathbf{J}_t \},\  \bar{\mathbf{J}}_t \in \{ \bar{\mathbf{J}}_t \}$}{
  $\bar{\mathbf{J}}_{t,0} = (\mathbf{J}_{t,0}-\mu_0)/\Sigma_0$;
  $\bar{\mathbf{J}}_{t,1} = (\mathbf{J}_{t,1}-\mu_1)/\Sigma_1$;\;
  $\bar{\mathbf{J}}_{t,2} = (\mathbf{J}_{t,2}-\mu_2)/\Sigma_2$;
  $\bar{\mathbf{J}}_{t,3} = (\mathbf{J}_{t,3}-\mu_3)/\Sigma_3$;\;
  $\bar{\mathbf{J}}_{t,4} = (\mathbf{J}_{t,4}-\mu_4)/\Sigma_4$;
  $\bar{\mathbf{J}}_{t,5} = (\mathbf{J}_{t,5}-\mu_5)/\Sigma_5$;\;
  }

  Divide $\{ \mathbf{J}_t \}$ into groups $\{\mathbf{g}_k\}$ using $\{ \bar{\mathbf{J}}_t\}$ as feature with voxel size of $D_{v}$;\;
  
  \While{ $S^{O}_{|\mathcalbf{L}}| < N^{O} $}{
      $\mathcalbf{G} \gets $ subset obtained by random sampling $\lambda \mathrm{log}(1/\epsilon)$ groups from $\{\mathbf{g}_k\}$;\;
      $\mathcalbf{R} = {\{\mathbf{J}}_j\} \gets $ subset obtained the first element in each sampled group $\mathcalbf{G}$ with ${\mathbf{J}}_j \in \mathbf{g}_k$;\;
      \ForEach{${\mathbf{J}}_j \in \mathcalbf{R}  $}{
      Compute $\varLambda_j = \mathbf{J}_j^{\top}\mathbf{J}_j$;\;
      }
      $j^{*}\gets \mathrm{argmax}_{j\in \mathcalbf{R}} \lambda_{min}[\varLambda(S^{O}_{\mathcalbf{L}})+\varLambda_j] $;\;
      $\varLambda(S^{O}_{\mathcalbf{L}}) \gets \varLambda(S^{O}_{\mathcalbf{L}})+\varLambda_j$;\;
      $S^{O}_{\mathcalbf{L}} \gets S^{O}_{\mathcalbf{L}} \bigcup r_{j^{*}}^{\mathcal{L}}$; $\mathbf{g}_k \gets \mathbf{g}_k \backslash  \mathbf{J}_{j^*}$. \;
      
  }
  \caption{Group-Based Stochastic-Greedy for Feature Selection Considering Optimality \label{algo:Group-Based Stochastic-Greedy}}
\end{algorithm}

Eq. \eqref{eq_worst_bound} elicits a lower bound for the stochastic-greedy heuristic results. In practice, the stochastic-greedy heuristics will achieve better results than the lower bound. 

\textbf{Theorem. 3} \citep{nemhauser1978analysis} The time complexity of the stochastic-greedy heuristic is $O(|S_{\mathcalbf{L}}|\mathrm{log}(1/\epsilon))$, which is related to $|S_{\mathcalbf{L}}|$ but independent of $N^{O}$.

\textbf{Group-based stochastic-greedy solver}: As the candidates with similar Jacobian have similar contributions to the objective function Eq. \eqref{eq_argmax}, we can speed up the stochastic-greedy search process by analyzing the similarity of Jacobians and group the candidate correspondences to avoid sampling similar candidate correspondences in each stochastic step as shown in Algorithm \ref{algo:Group-Based Stochastic-Greedy}. We treat each Jacobian vector $\textbf{J}_t$ as a 6-dimensional feature (3 for quaternion and 3 for translation). Since the numerical range of the quaternion part and translation part of $\textbf{J}_t$ differ significantly, we regularize each dimension of $\textbf{J}_t$ to get $\bar{\textbf{J}}_t$ (Line 2-7 in  Algorithm \ref{algo:Group-Based Stochastic-Greedy}). Second, we segment $S_{\mathcalbf{L}}$ into groups considering the regularized Jacobian via a high-dimensional voxelization with a voxel size $D_{v}$ to get $\lambda N^{O}$ groups as illustrated in Fig.\ref{fig_sampling}. The above grouping processing's complexity is $O(|S_{\mathcalbf{L}}|)$ by saving the voxel indexes in the hash table. At last, in each greedy searching step, we randomly get $\lambda \mathrm{log}(1/\epsilon)$ candidates from each group and select one best candidate (other than $|S_{\mathcalbf{L}}|/N^{O}\mathrm{log}(1/\epsilon) $ candidates in the original stochastic-greedy search process). The iterative process stops until it gets $N^{O}$ correspondences (Line 9-18 in Algorithm \ref{algo:Group-Based Stochastic-Greedy}). 

\textbf{Lazy counting strategy}: To determine the voxel size $D_{v}$, a lazy counting strategy (LCS) is proposed. LCS first downsamples the current candidates with farthest point sampling \citep{qi2017pointnet++} to the size of $\lambda N^{O}$. Then the mean distance between the remaining candidates is calculated to get $D_{v}$. As the feature distribution in the environment does not dramatically change every frame. LCS is only triggered to update $D_{v}$ when the size of the grouping results differ a lot ($20\%$ in practice) from $\lambda N^{O}$.

\subsection{LiDAR keyframe selection and maintenance \label{4.4}}

We select a keyframe when the system's translation or orientation changes over a certain threshold. Thus, the keyframe database consists of keyframes distributed in time or space. The points in the keyframe are added to the local map stored in ikdtree \citep{xu2022fast}. Due to the length limitation, this paper only focuses on LIO without loop closure.

\section{Experimental Validation of the System \label{section_exp}}
 We demonstrate the performance of HCTO via the public wearable sensing dataset (WHU-Helmet \citep{li2023whu}) and in-house wearable sensing datasets. We compare our method with three other state-of-the-art (SOTA) methods. First is Fast-Lio2 \citep{xu2022fast}, which uses filter-based state estimation with the ikdtree managing the whole map points in an incremental way. Second is DLIO \citep{chen2022dlio}, which is a direct LIO using continuous-time motion undistortion. Third is SLICT, which estimates the system states using CTO. Fourth is CLINS \citep{lv2021clins}, which enables fuse high-frequency and asynchronous sensor data effectively. Video recordings of experiments could be found on the project page of HCTO: \href{https://github.com/kafeiyin00/HCTO}{https://github.com/kafeiyin00/HCTO}.

\subsection{Implementation}

We implemented the proposed HCTO in C++ and Robot Operating System (ROS). The order of the B-Spline is set to 4. The time segment length $\Delta t$ for the B-Spline is set to 0.05. The decay factor $\epsilon$ for the stochastic-greedy is set to 0.1. The size of the selected correspondences $N^{O}$ is set to 500. Grouping factor $\lambda$ is set to 2 to get 1000 ($\lambda N^{O}$) correspondence groups in each LiDAR frame. 

\subsection{Public WHU-Helmet Dataset}

\begin{figure}[h]
\centering
\includegraphics[width=0.8\textwidth]{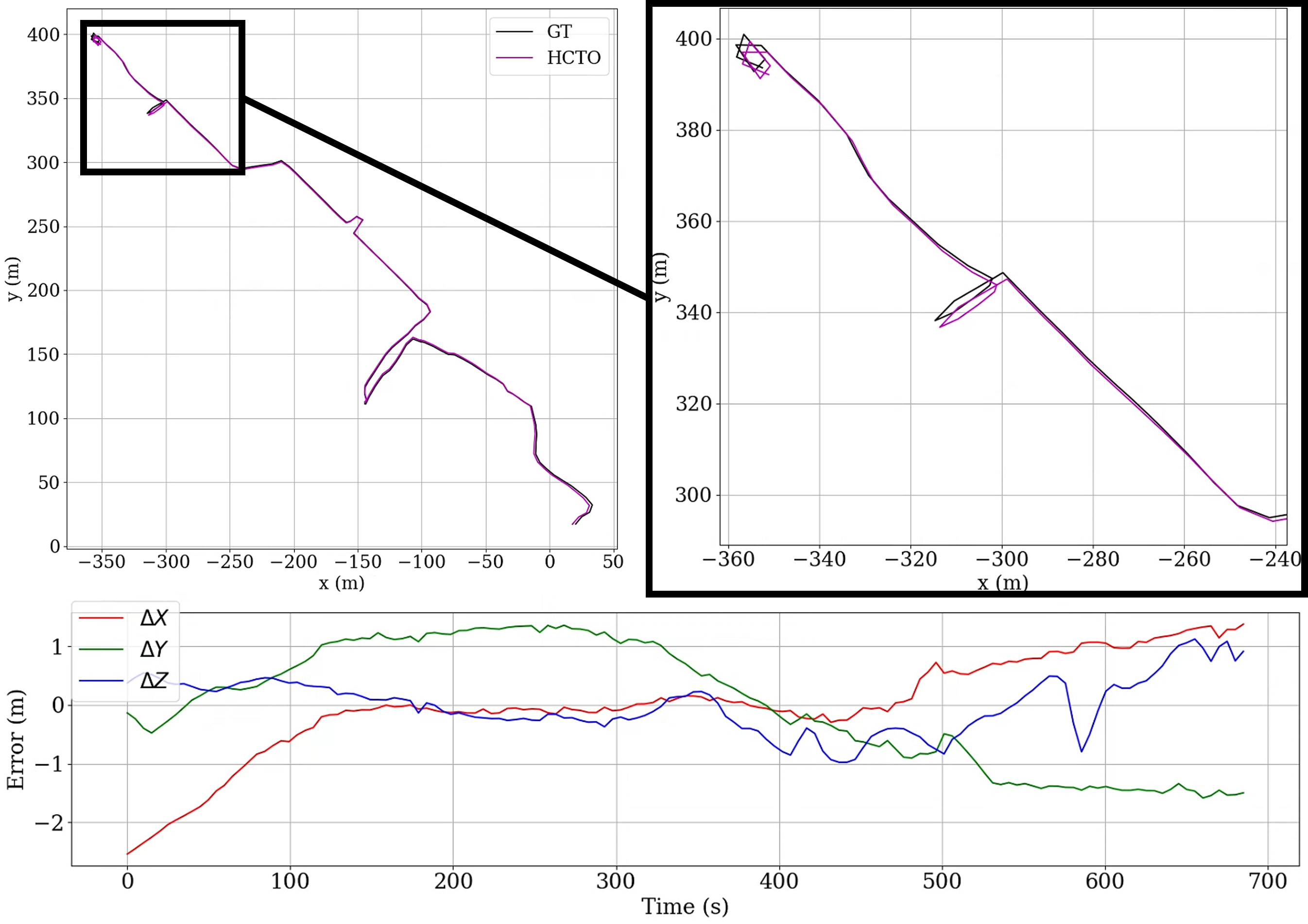}
\caption{ Localization error analysis of the seq-subway-station in WHU-Helmet. The
first row is the trajectory of HCTO. The second row is the
localization errors in different axes of HCTO. }\label{fig_WHU_HELMET_ATE}
\end{figure}

\begin{figure}[!t]
\centering
\includegraphics[width=0.8\textwidth]{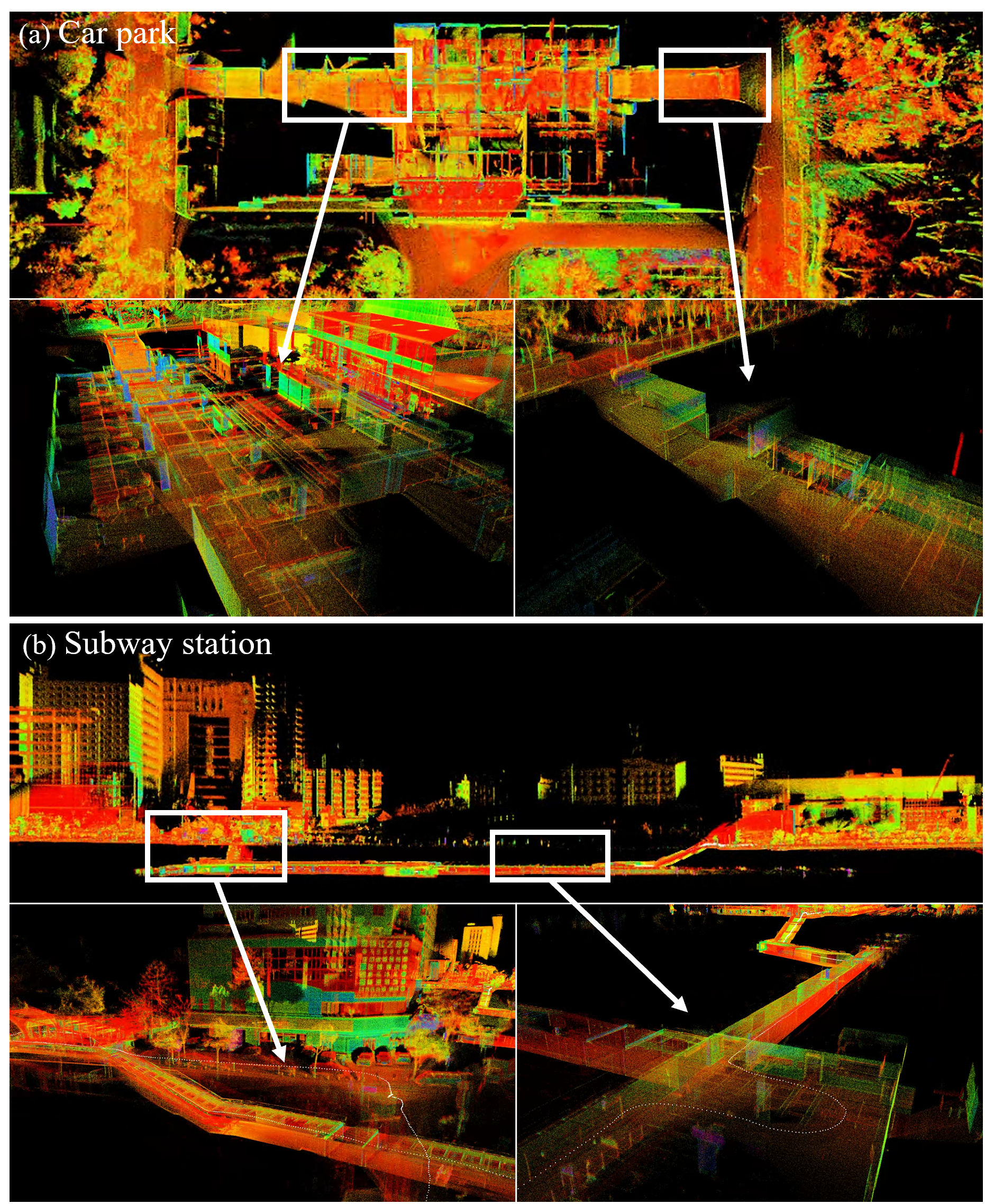}
\caption{ The point cloud maps constructed by the proposed HCTO in WHU-Helmet Datasets. (a) Car park sequence; (b) Subway station sequence. }\label{fig_WHU_HELMET_dataset}
\end{figure}

The WHU-Helmet dataset was also collected by the helmet-based wearable system. However, the main difference in the hardware configuration between WHU-Helmet and the in-house built one is the LiDAR scanner. The Livox Avia is used in WHU-Helmet, which has a longer observation range but with a limited sight view ($ 70.4^{\circ} \times 77.2^{\circ}$) compared to the Livox Mid360 (with a sight view of $ 360^{\circ} \times 59^{\circ}$) used in the proposed system. The limited view also poses a challenge for LIO systems in indoor environments. We selected two sequences, the car park and subway station scenes, from the WHU-Helmet dataset \citep{li2023whu} to evaluate the performance of the proposed method. Both car parks and subway stations are common scenes, which needed prior maps for the robot-based delivery. We use Absolute Translation Error (ATE) \citep{grupp2017evo} as the metric to evaluate the performance of different methods and list the results in Table \ref{tab:ATE_WHU_HELMET}. As the two scenarios contain too many degenerated scenes like the corridor and elevator passage, most of the existing methods diverged. The localization error of seq-subway-station in WHU-Helmet for the proposed HCTO is plotted in Fig.\ref{fig_WHU_HELMET_ATE}. The proposed HCTO still achieved good performance in these degenerated scenes.

The point cloud maps constructed by the proposed HCTO for the subway station and car park are illustrated in Fig.\ref{fig_WHU_HELMET_dataset}. From visual inspection, the point cloud maps constructed by the proposed HCTO achieved high accuracy and did not diverge in the narrow corridors and elevator passage. The experimental results demonstrated the potential for the proposed method in the prior map construction in car parks and subway stations.

\begin{table*}[]
\caption{ATE of HCTO and other methods on public WHU-Helmet Dataset (Unit [m]). The best results are in \textbf{BOLD}, second best results are \underline{underlined}, x denotes diverge. \label{tab:ATE_WHU_HELMET}}
\centering
\begin{threeparttable}
\scalebox{0.8}{
\begin{tabular}{lccccc}
\hline \hline
Dataset & Fast-Lio2 & DLIO & SLICT & CLINS & HCTO (Ours) \\ \hline
seq-car-park ($\sim0.6 km$, \textbf{W})  & x & x & x & \underline{1.16} & \textbf{0.504}       \\
seq-subway-station ($\sim0.8 km$, \textbf{W}) &  \underline{3.55} & x & x & 3.94 & \textbf{1.410}      \\ \hline\hline
\end{tabular}
}
\begin{tablenotes} \tiny
\item \textbf{W} for Walking  
\end{tablenotes}
\end{threeparttable}
\end{table*}

\subsection{In-House NTU-Campus Dataset}
\begin{figure}[h]
\centering
\includegraphics[width=0.8\textwidth]{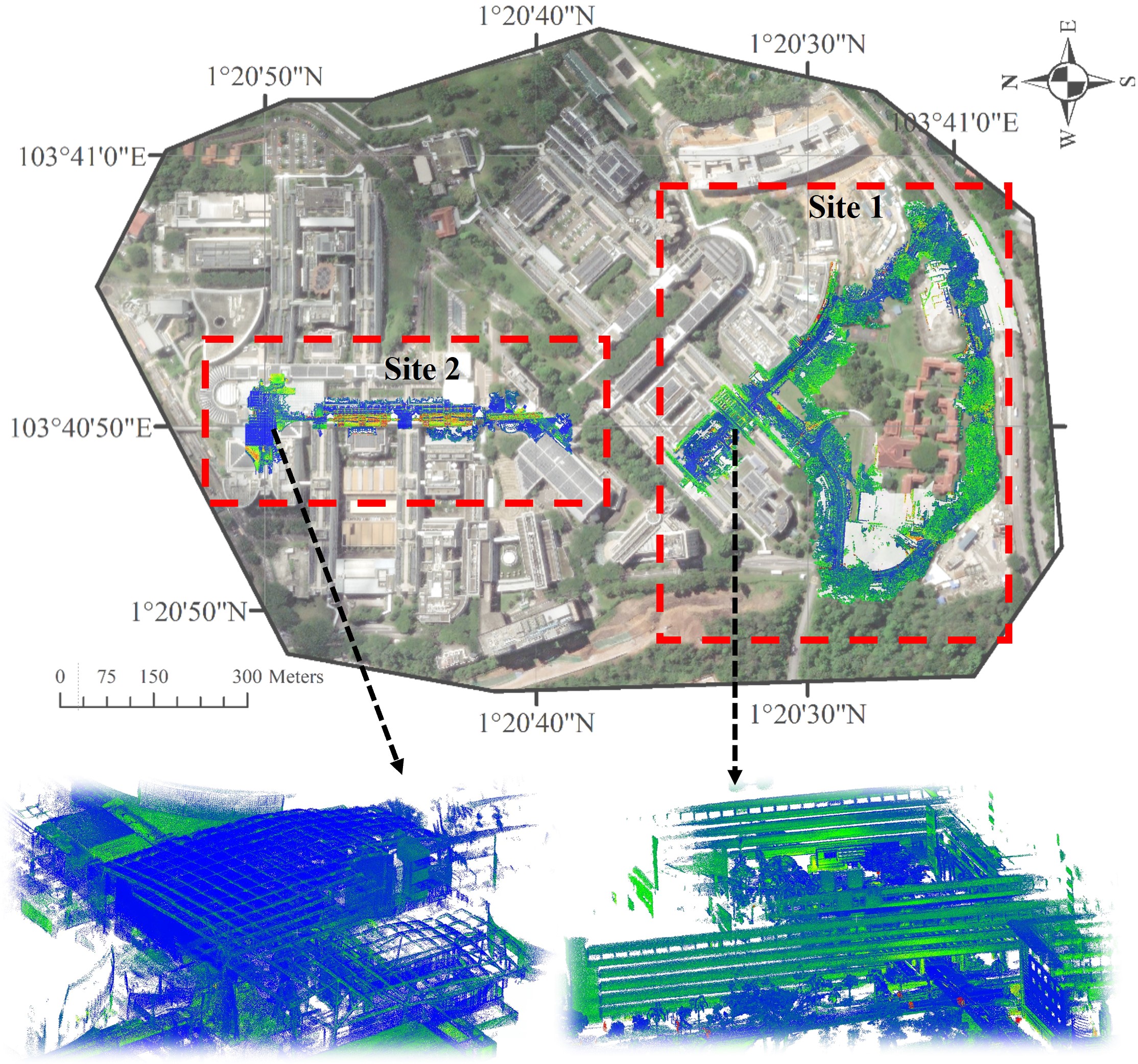}
\caption{The NTU-Campus Datasets. Site 1 is an open roadside environment. Site 2 is a multi-level indoor environment. The intensity rendered point clouds are overlapped on the satellite image of NTU. }\label{fig_NTU_campus_dataset}
\end{figure}

\begin{figure}[h]
\centering
\includegraphics[width=\textwidth]{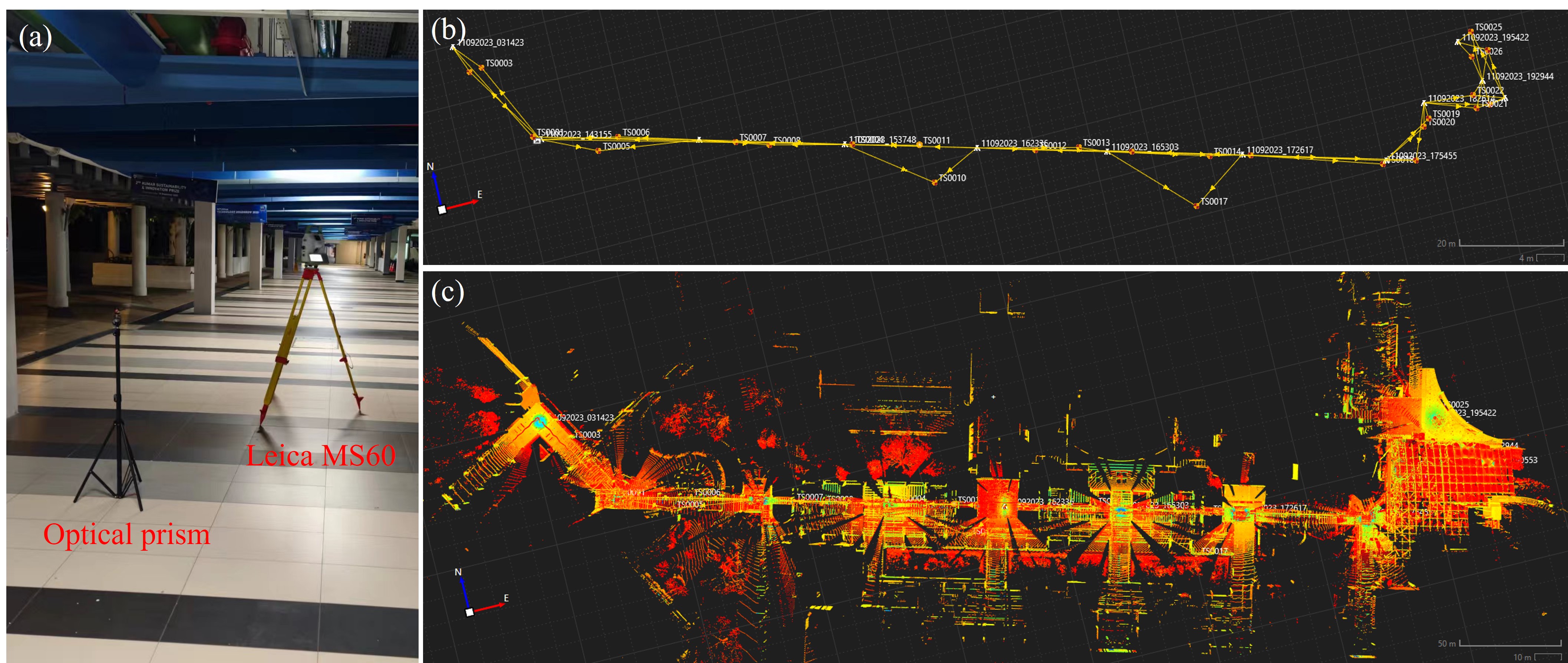}
\caption{Ground truth collection using Terrestrial Laser Scanning (TLS) of site 2 in the NTU campus dataset. (a) Setup of control points in the environment using optical prisms. (b) The geodetic network for TLS data registration. (c) The obtained centimeter-level ground truth map. }\label{fig:gt_collection_leica}
\end{figure}

\begin{figure}[h]
    \centering
    \includegraphics[width=0.96\textwidth]{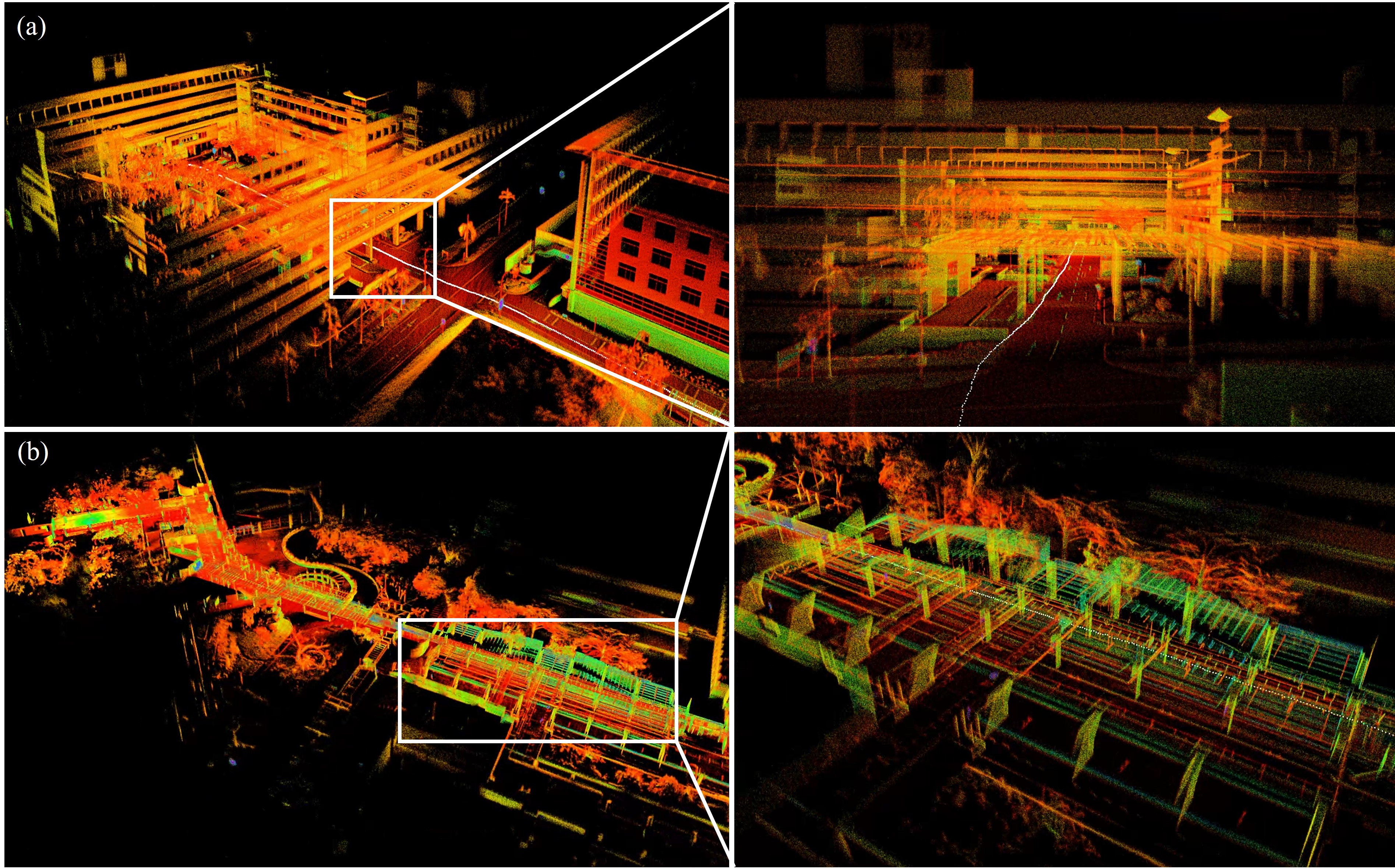}
    \caption{ Visualization of the point clouds generated by HCTO in NTU campus. (a) Side view of a random LiDAR frame in Site 1. (b) Side view of a random LiDAR frame Site 2.}\label{fig_NTU_campus_view}
\end{figure}

\begin{figure}[h]
    \centering
    \includegraphics[width=0.96\textwidth]{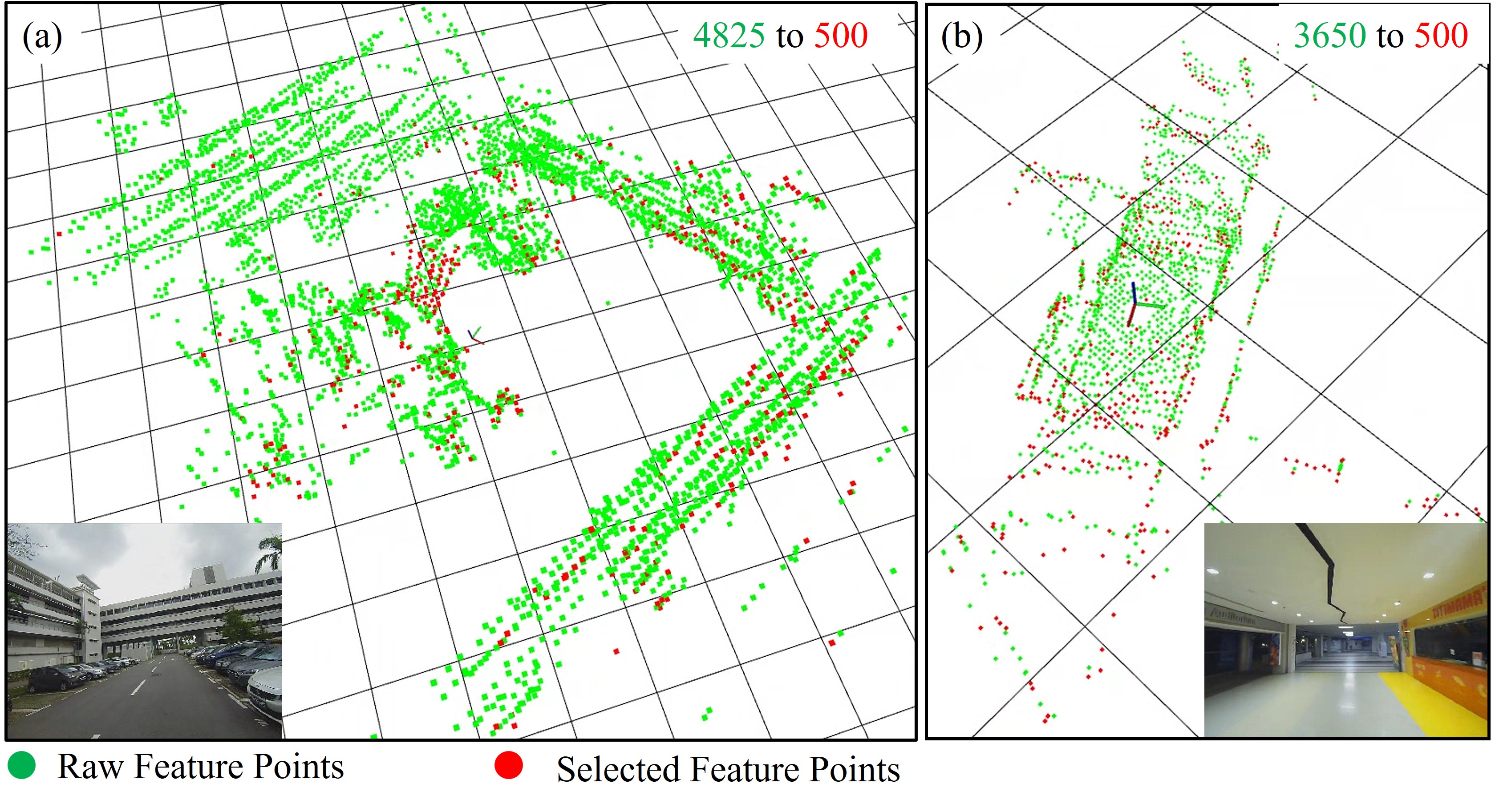}
    \caption{ Visualization of feature selection in NTU campus dataset. (a) Side view of a random frame in Site 1. (b) Side view of a random frame in Site 2.}\label{fig_feature_selection_NTU_campus}
\end{figure}

The NTU-Campus dataset, containing two study sites in Nanyang Technological University as shown in Fig.\ref{fig_NTU_campus_dataset}, was collected using the in-house built compact helmet system. Site 1 is an open-road environment that covers an area of about $500m \times 300m$. Site 2 is a multi-level indoor environment that covers an area of $400m \times 100m$. The two study sites represent common environments for robot delivery. The ground truth is constructed in a similar manner to the Newer College Dataset \citep{ramezani2020newer}. We utilized the centimeter-level terrestrial laser scanner (Leica MS60) to obtain the prior map at the study site. To guarantee the accuracy of the prior map, we first evenly set a lot of optical prisms as the control points in the environment as shown in Fig.\ref{fig:gt_collection_leica} (a). Then a standard geodetic network adjustment \citep{teunissen1985quality} is applied for the terrestrial laser scanning registration as shown in Fig.\ref{fig:gt_collection_leica} (b). The ground truth centimeter-level point clouds map of Site 2 is shown in Fig.\ref{fig:gt_collection_leica} (c). Each LiDAR scan from the wearable sensing system was registered to the prior map to obtain the ground truth pose.

\begin{figure}[h]
\centering
\includegraphics[width=0.86\textwidth]{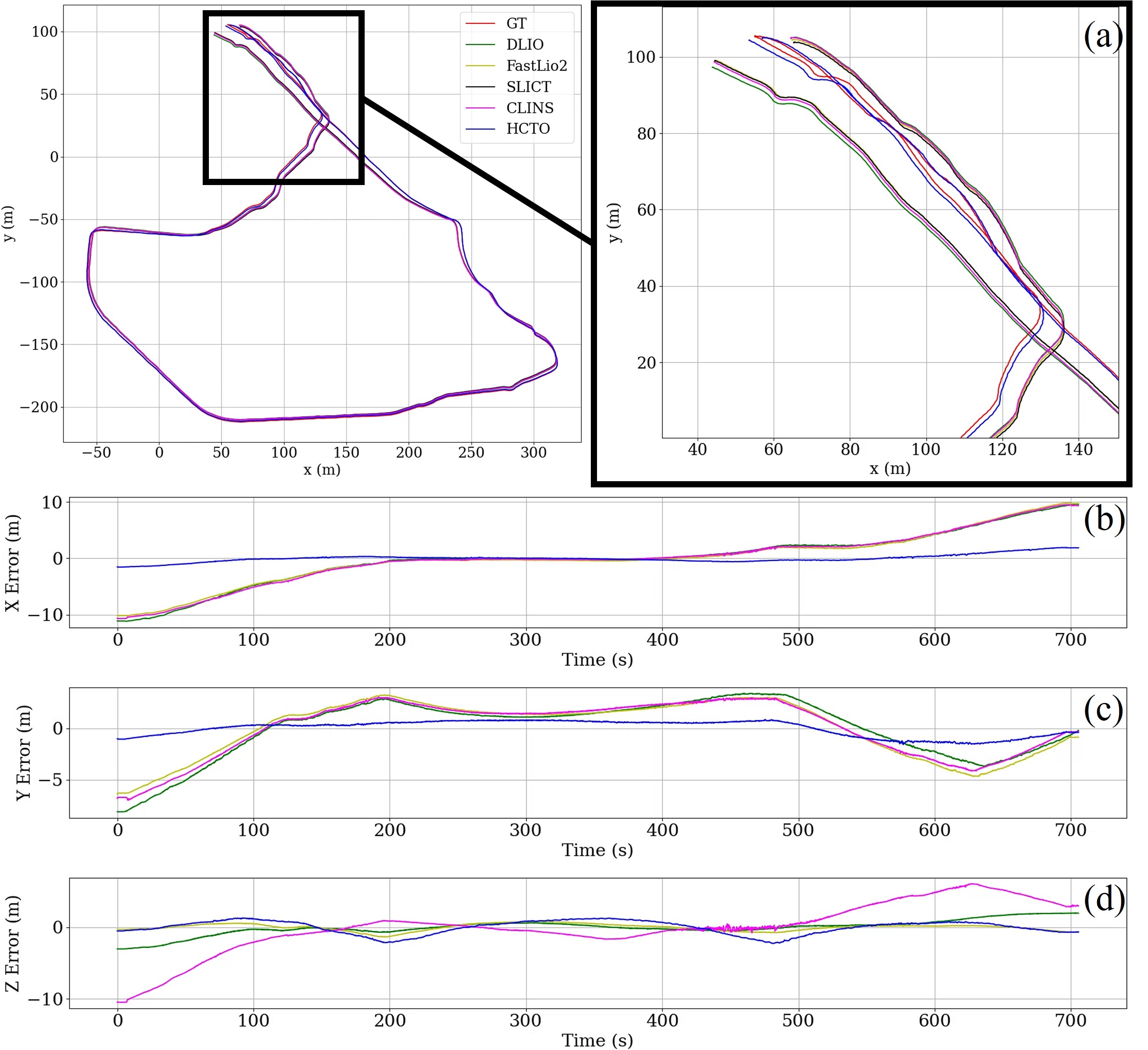}
\caption{Localization error analysis of the seq-02 in NTU-Campus site 1. (a) The trajectories from different methods. (b),(c) and (d) The localization errors in different axes.}
\label{fig_NTU_Campus_seqs_01}
\end{figure}

In site 1, the operator collected three data sequences (seq-01, seq-02, and seq-03) under varying degrees of motion: walking (about $1.5 m/s$), running (about $2.5 m/s$), and a combination of walking and running, respectively. The trajectory length of each sequence is about $1500 m$. When the operator was running, the data sequence contained more vibration than the walking sequence. Visualization of the point clouds generated by HCTO in Site 1 is shown in Fig.\ref{fig_NTU_campus_view} (a). The feature selection result of a random LiDAR frame is visualized in Fig.\ref{fig_feature_selection_NTU_campus} (a). It could be seen that 500 good features are kept from 4825 features, which guarantees real-time performance.

Similar to previous experiments, we used ATE as the metric to evaluate the accuracy of different methods. The trajectories and localization errors from different methods are plotted in Fig.\ref{fig_NTU_Campus_seqs_01}. The final results are listed in Table \ref{tab:ATE_NTU_CAMPUS}. As the operator moved more gently in the first sequence, most methods achieved better results compared to the other two sequences. As the operator ran to collect the seq-02, all the methods suffered from the vibration. Table \ref{tab:ATE_NTU_CAMPUS} also illustrated that the proposed HCTO achieved the best performance for all three sequences in site 1.

\begin{figure}[!t]
\centering
\includegraphics[width=0.86\textwidth]{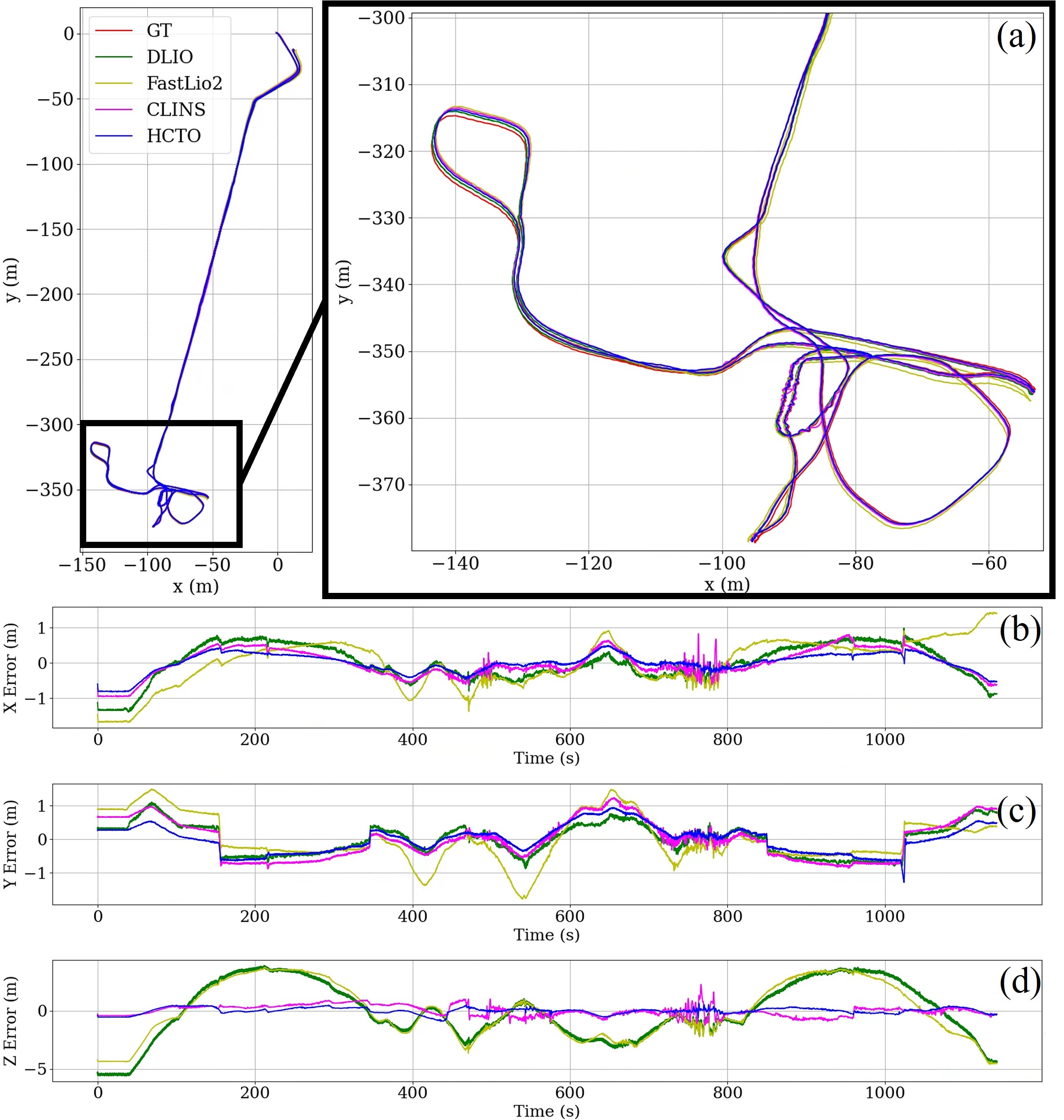}
\caption{Localization error analysis of the seq-01 in NTU-Campus site 2. (a) The trajectories from different methods. (b),(c) and (d) The localization errors in different axes.}
\label{fig_NTU_Campus_seqs_02}
\end{figure}

In site 2, the operator walked (about $1.5 m/s$) to collect data sequence (seq-01) in a very complicated indoor environment, the Nanyang Auditorium in NTU, which contains multiple levels and degenerated corridor scenes. Visualization of the point clouds generated by HCTO in Site 1 is shown in Fig.\ref{fig_NTU_campus_view} (b). The feature selection result of a random LiDAR frame is visualized in Fig.\ref{fig_feature_selection_NTU_campus} (b).  The trajectories and localization errors from different methods are plotted in Fig.\ref{fig_NTU_Campus_seqs_02}. From the results, it can be found that the proposed HCTO achieved trajectory accuracy, especially in the Z direction. The APE of the trajectory errors were also listed in Table \ref{tab:ATE_NTU_CAMPUS}. Due to the narrow corridors in the environment, SLICT diverged and did not get a valid result. 

\begin{figure}[!t]
\centering
\includegraphics[width=0.8\textwidth]{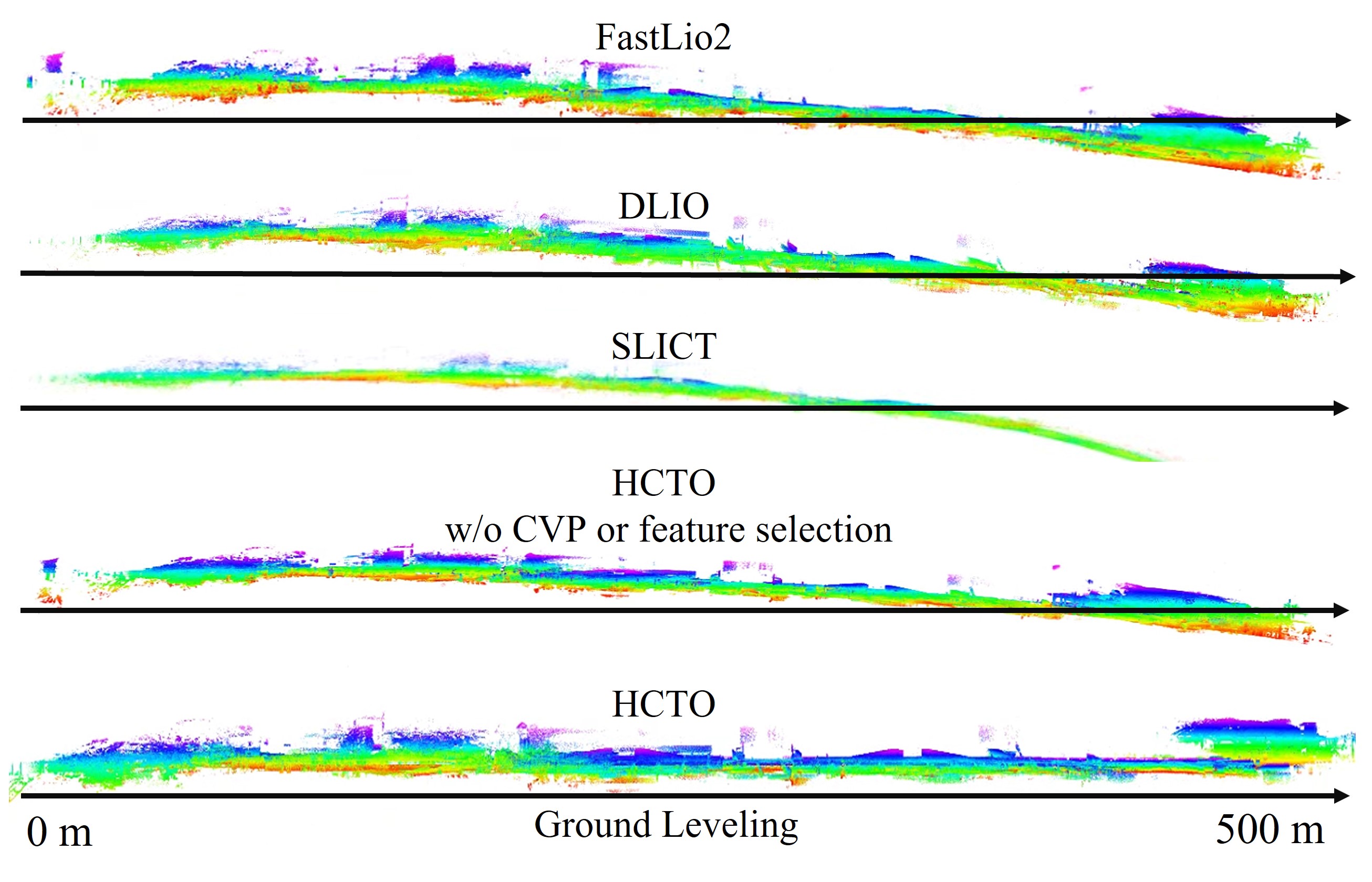}
\caption{Visual comparison of the different methods in NTU Campus site 2. The point clouds are rendered by height to have better visualization of the drift in attitude and z direction. }
\label{fig_NTU_Campus_site2_visual_compare}
\end{figure}

In site 2, we also evaluated the quality of different methods by visual inspection of the walls and ground constructed in the point map. We have checked the ground and the walls in the real scenes were level and coincident with the gravity direction, respectively. The visual comparison between point clouds generated by different methods is illustrated in Fig.\ref{fig_NTU_Campus_site2_visual_compare}. From the visual inspection, only the point clouds generated by the proposed HCTO achieved the leveling of the ground and straight-up walls, which is vital for a prior map used for robot delivery. Otherwise, the robot can not achieve the correct goal in a bent prior map. We also disabled the CVP factor and feature selection model in HCTO and output the point clouds, which also have little attitude drift from the visual inspection. The main reason for the bent map constructed by the LIO methods is that the unevenly distributed correspondences and degenerated scenes made the LIO systems converge to the local minimal solution and lose the observation of attitude, especially under the high vibration. However, the proposed HCTO used the CVP factor to maintain the correct attitude and select good features to balance the unevenly distributed correspondences, thus achieving the best performance. It also should be noted that due to the field of view limitation of the MiD360 on the helmet, the ground may not be observed in the LiDAR frame. Thus, the strategy using ground points to restrict the drift \citep{shan2018lego} can be difficult to be directly applied to the helmet system.

\begin{table*}[]
\caption{ATE of HCTO and other methods on in-house NTU-Campus Site (Unit [m]). The best results are in \textbf{BOLD}, second best results are \underline{underlined}. X denotes diverge.\label{tab:ATE_NTU_CAMPUS}}
\centering
\begin{threeparttable}
\scalebox{0.8}{
\begin{tabular}{lccccc}
\hline \hline
Dataset & Fast-Lio2 & DLIO & SLICT & CLINS & HCTO (Ours) \\ \hline
site1-seq-01 ($\sim1.5 km$, \textbf{W})  &  5.103 & 5.437 & \underline{4.383} & 5.462 & \textbf{2.059}       \\
site1-seq-02 ($\sim1.5 km$, \textbf{R}) &  \underline{5.262} & 5.503 & 5.547 & 6.337 & \textbf{3.446}      \\
site1-seq-03 ($\sim1.5 km$, \textbf{W\&R}) & 5.228 & 5.759 & 4.452 & \underline{3.406} & \textbf{3.105}         \\
site2-seq-01 ($\sim1.0 km$, \textbf{W}) & 2.528 & 2.548 & x & \underline{1.823}  &  \textbf{0.551} \\\hline\hline
\end{tabular}
}
\begin{tablenotes} \tiny
\item \textbf{W} for walking; \textbf{R} for running; \textbf{W\&R} for a combination of walking and running  
\end{tablenotes}
\end{threeparttable}
\end{table*}

\subsection{Degenerated Scenes of Multi-level Apartment}

\begin{figure*}[!t]
    \centering
    \includegraphics[width=\textwidth]{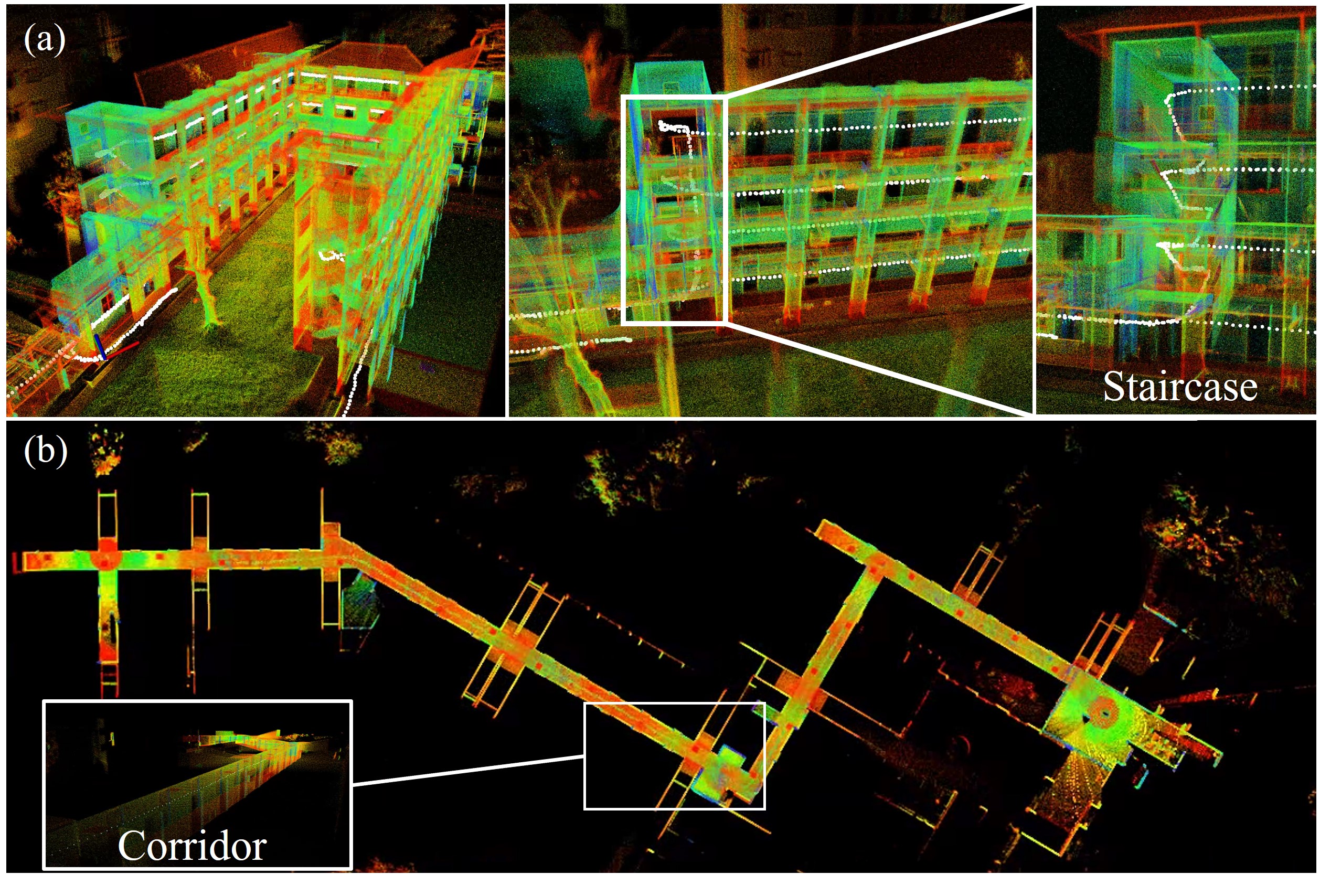}
    \caption{Prior map constructed using the proposed HCTO. (a) Outside view of a four-floor apartment. The operator ran to collect data floor by floor going through the staircase, which is a degenerated scene. (b) Inside view of the apartment. The long corridor of each floor is also challenging for the LIO system.}
    \label{fig_apartment_points}
\end{figure*}

Apartments in hospitals and nursing homes are typical scenes that need robot delivery. The apartments usually contain a lot of degenerated scenes like staircases and long corridors, which pose challenges especially when the platform is under shaking. We validated our system in a multi-level apartment from outside to indoor as shown in Fig.\ref{fig_apartment_points}. The operator ran to collect data from the first floor to the top floor through the staircase as shown in Fig.\ref{fig_apartment_points} (a). As for the inside of the apartment, the operator walked through the long corridor as shown in Fig.\ref{fig_apartment_points} (b). Most existing LIO systems diverged when the operator ran into the staircase or walked in the corridor. However, the proposed HCTO still achieved pretty good results. It should be noted that we only took about 15 minutes to construct the prior map for the multi-level apartment using our wearable system, which is much more efficient and cost-effective than the standard survey solution \citep{dong2020registration}.

\subsection{Time performance of HCTO}

\begin{figure}[h]
    \centering
    \includegraphics[width=0.96\textwidth]{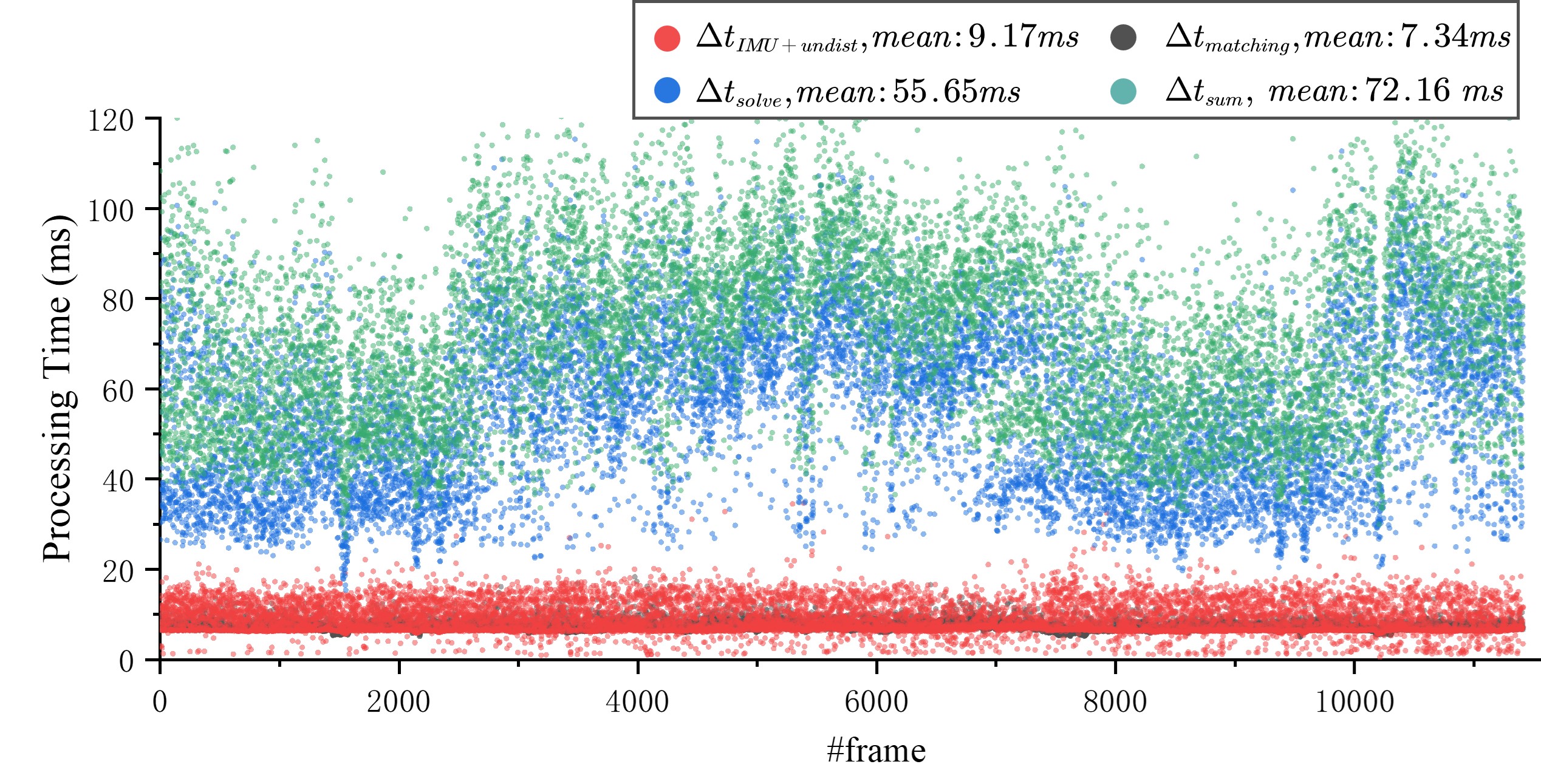}
    \caption{The time performance analysis in NTU-Campus site 2.}
    \label{fig_time_analysis}
\end{figure}

We analyze the computation load of HCTO on the NTU-Campus site 2 in Fig.\ref{fig_time_analysis}. The experiment was run on a computer with an Intel Core I9-I2900 CPU. On average, it took 72.16 ms ($\Delta t_{sum}$) to complete one LiDAR frame procession, in which the pure IMU processing and LiDAR undistortion took 9.17 ms ($\Delta t_{IMU+undist}$), the LiDAR feature association took  7.34 ms ($\Delta t_{matching}$), the feature selection and state estimation took 55.65 ms ($\Delta t_{solve}$), which guaranteed the real-time performance (10 Hz) of the proposed system. We recorded the memory usage and time performance for all the sequences, which can be found on the project page of HCTO.

\subsection{Ablation study of HCTO-based LIO}

\begin{table*}[h]
\caption{ATE in different helmet-based sequences with ground truth (Unit [m]). The results with increased ATE compared with HCTO are in \TG{GREEN}. The results with decreased ATE compared with HCTO are in \TR{RED}. x denotes diverge. \label{tab:ATE_ABLATION}}
\centering
\begin{threeparttable}
\scalebox{0.6}{
\begin{tabular}{llcccc}
\hline \hline
Dataset & Sequence & HCTO (w/o \textbf{HB}) & HCTO (w/o \textbf{CVP}) & HCTO (w/o \textbf{FS})  & HCTO\\ \hline
\multirow{2}{*}{WHU-Helmet}&seq-car-park ($\sim0.6 km$, \textbf{W})  &  x (\TG{+$\infty$})  & 0.801 (\TG{+0.297})  &  0.688 (\TG{+0.113}) &0.504       \\
&seq-subway-station ($\sim0.8 km$, \textbf{W}) & 1.832 (\TG{+0.422})   & 1.644 (\TG{+0.234})  & 2.751 (\TG{+1.341}) & 1.410      \\ \hline
\multirow{3}{*}{NTU-Campus}&site1-seq-01 ($\sim1.5 km$, \textbf{W})  &  5.575 (\TG{+3.516})  & 2.819 (\TG{+0.760})  &  2.013 (\TR{-0.046}) & 2.059       \\
&site1-seq-02 ($\sim1.5 km$, \textbf{R}) & x (\TG{+$\infty$}) & 3.580 (\TG{+0.134}) &  3.312 (\TR{-0.134})  & 3.446      \\
&site1-seq-03 ($\sim1.5 km$, \textbf{W\&R}) & 4.603 (\TG{+1.498}) & 3.315 (\TG{+0.210}) &  2.893 (\TR{-0.212})  & 3.105         \\ \hline  
\hline
\end{tabular}
}
\begin{tablenotes} \tiny
\item \textbf{W} for walking; \textbf{R} for running; \textbf{W\&R} for a combination of walking and running  
\item \textbf{HB} for hybrid IMU factors; \textbf{CVP} for constant velocity factor; \textbf{FS} for feature selection.
\end{tablenotes}
\end{threeparttable}
\end{table*}

To investigate the contribution of each design feature in HCTO's performance, we set different configurations of HCTO and compare the results on the helmet-based dataset with ground truth trajectories, and listed the results in Table.\ref{tab:ATE_ABLATION}. We disabled the hybrid IMU factors and use the original IMU factors (same as the LFP) for all motion states, whose results were listed in the first column. We disabled the CVP factors and listed the results in the second column. At last, we disabled the feature selection module and retained all the correspondences for optimization without considering the time performance, whose results are listed in the third column. 

The first column of Table. \ref{tab:ATE_ABLATION} illustrates that using the hybrid IMU factors could suppress the vibration effect. The high-frequency part can not be fitted by the B-Spline trajectory, which may cause a large drift or even divergence. The second column of Table. \ref{tab:ATE_ABLATION} illustrates that CVP factor could limit the drift. From the results of NTU-Campus dataset, the CVP factor got higher performance improvement when the operator was walking. Because the CVP period may not exist when the operator was running. The third column of Table. \ref{tab:ATE_ABLATION} illustrated that the feature selection module achieved performance improvement in degenerated scenes in the WHU-Helmet dataset. It should be noted that HCTO (w/o \textbf{FS}) could not perform in real-time as it retained all the correspondences.

In summary, we conclude that the hybrid IMU factor is effective in ensuring robustness towards high vibration. The CVP factor can noticeably reduce the drift, and the feature selection scheme can be very effective in degenerative cases. In cases where feature selection reduces the accuracy, it is only a minor decrease that can be justified by the significant improvement in real-time performance.

\section{Conclusion \label{section_conclusion}}

This paper proposed HCTO, an optimality-aware LIO with hybrid continuous-time optimization achieving significant accuracy and robustness for compact wearable mapping systems under high vibration conditions and degenerated environments. Extensive experiments on head-mounted WMS datasets were conducted to evaluate the performance of our system. The results demonstrated significant advantages of HCTO over state-of-the-art methods on a helmet-based WMS. Compared with existing SOTA LIO methods, the gains of HCTO in robustness and accuracy are due to constructing different types of IMU factors for different human motion states and optimal design for the LiDAR correspondences. The proposed compact wearable mapping system is used to collect prior maps for robot-based "last-mile" delivery in complex indoor environments.

There remain many possibilities for the extension of HCTO. For one, the motion-based factor for state estimation is simple and not fully explored. More efficient motion-based factors and activity-based loop information can be made by learning-based strategies \citep{zhou2023asl, wang2022a2dio}. Moreover, we will explore the possibility of integrating HCTO into other platforms (e.g., UAV, robot dog) by utilizing motion dynamics \citep{bauersfeld2021neurobem, buchanan2022learning}.

\section{Acknowledgment}
	
This research is supported by the National Research Foundation, Singapore under its Medium Sized Center for Advanced Robotics Technology Innovation.

\section {Declaration of Interests}
    The authors declare that they have no known competing financial interests or personal relationships that could have appeared to influence the work reported in this paper.

\bibliographystyle{elsarticle-harv} 
\bibliography{cas-refs}

\end{document}